\documentclass[letterpaper]{article} 
\usepackage{aaai25}

\usepackage{times}  
\usepackage{helvet}  
\usepackage{courier}  
\usepackage[hyphens]{url}  
\usepackage{graphicx} 
\usepackage{natbib}  
\usepackage{caption} 
\usepackage{algorithm}
\usepackage{algorithmic}
\usepackage{amsmath}
\usepackage{amssymb}
\usepackage{booktabs}
\usepackage{multirow}
\usepackage[table]{xcolor}
\usepackage{subcaption}
\usepackage{tabularx}
\usepackage{tikz}
\usepackage{adjustbox}
\usepackage{xcolor}
\usepackage{pifont}
\usepackage{array}
\usepackage{colortbl}
\usepackage{pgfplots}
\usepackage{romannum}

\newcommand{\x}{×}
\usepackage{pifont}  
\newcommand{\xmark}{\ding{55}}  
\newcommand{\cmark}{\ding{51}}  

\usepackage[capitalize]{cleveref}
\crefname{section}{Sec.}{Secs.}
\Crefname{section}{Section}{Sections}
\Crefname{table}{Table}{Tables}
\crefname{table}{Tab.}{Tabs.}

\newcommand{\eg}{\emph{e.g.}}
\newcommand{\ie}{\emph{i.e.}}

\definecolor{forestgreen}{RGB}{34,139,34}

\usepackage{newfloat}
\usepackage{listings}
\DeclareCaptionStyle{ruled}{labelfont=normalfont,labelsep=colon,strut=off} 
\floatstyle{ruled}
\newfloat{listing}{tb}{lst}{}
\floatname{listing}{Listing}
\title{Storyboard guided Alignment for Fine-grained Video Action Recognition}
\author{
    Enqi Liu\textsuperscript{\rm 1},
    Liyuan Pan\textsuperscript{\rm 1*},
    Yan Yang\textsuperscript{\rm 2},
    Yiran Zhong\textsuperscript{\rm 3},
    Zhijing Wu\textsuperscript{\rm 1},
    Xinxiao Wu\textsuperscript{\rm 1},
    Liu Liu\textsuperscript{\rm 4}
}
\affiliations{
    \textsuperscript{\rm 1}Beijing Institute of Technology,
    \textsuperscript{\rm 2}Australian National University\\
    \textsuperscript{\rm 3}Shanghai AI Lab,
    \textsuperscript{\rm 4}Huawei Technologies Ltd\\
    enqi.liu@bit.edu.cn
}

\usepackage{bibentry}

\begin{document}

\maketitle


\begin{abstract}

Fine-grained video action recognition can be conceptualized as a video-text matching problem. Previous approaches often rely on global video semantics to consolidate video embeddings, which can lead to misalignment in video-text pairs due to a lack of understanding of action semantics at an atomic granularity level. To tackle this challenge, we propose a multi-granularity framework based on two observations: (i) videos with different global semantics may share similar atomic actions or appearances, and (ii) atomic actions within a video can be momentary, slow, or even non-directly related to the global video semantics. Inspired by the concept of storyboarding, which disassembles a script into individual shots, we enhance global video semantics by generating fine-grained descriptions using a pre-trained large language model. These detailed descriptions capture common atomic actions depicted in videos. A filtering metric is proposed to select the descriptions that correspond to the atomic actions present in both the videos and the descriptions. By employing global semantics and fine-grained descriptions, we can identify key frames in videos and utilize them to aggregate embeddings, thereby making the embedding more accurate. Extensive experiments on various video action recognition datasets demonstrate superior performance of our proposed method in supervised, few-shot, and zero-shot settings.

\end{abstract}

\section{Introduction}

\begin{table}[t]
    \centering
    \captionsetup{type=figure}
    \begin{subtable}[t]{\linewidth} 
        \centering
        \vspace{-0.3cm}
        \begin{tikzpicture}
            \node (img) {\begin{tabular}{cccccc}
                \hspace{-0.59cm}\raisebox{-0.16cm}{
                \rotatebox{90}{\parbox[c][0.5cm][c]{1.5cm}{\centering\scalebox{0.8}{\small\text{Baking cookies}}}}}&
                \hspace{-0.49cm}\includegraphics[width=2.02cm, height=1.19cm]{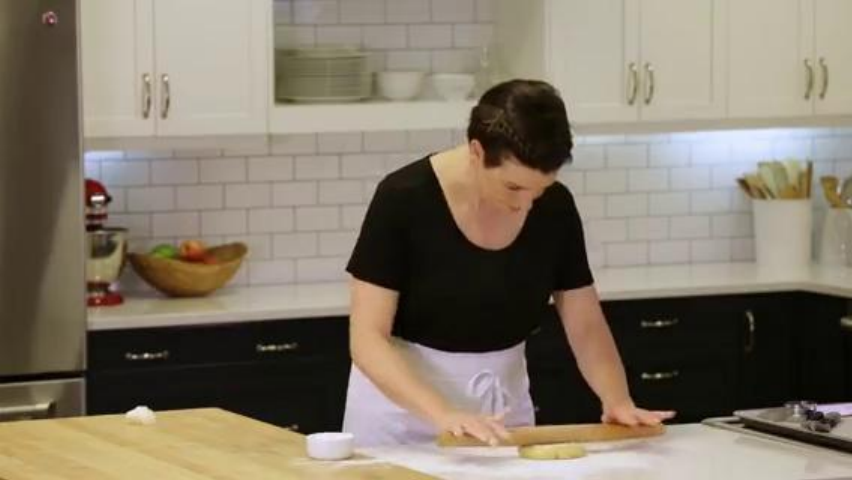} & 
                \hspace{-0.4cm}\includegraphics[width=2.02cm, height=1.19cm]{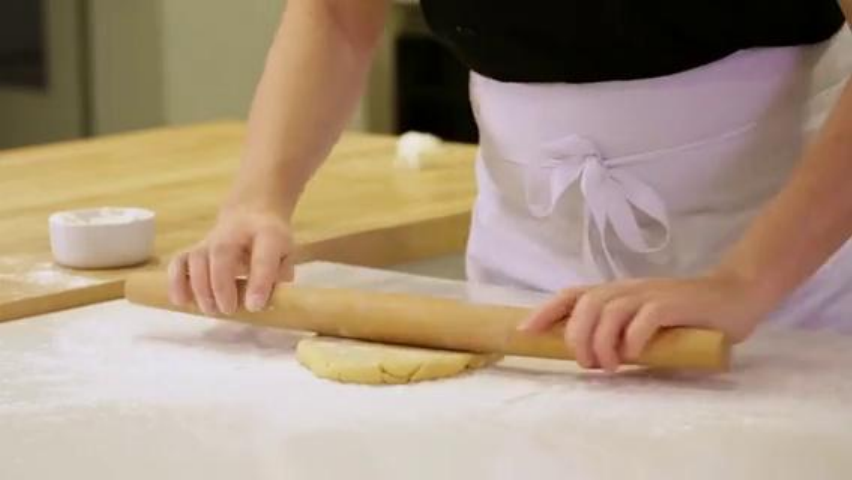} & 
                \hspace{-0.4cm}\includegraphics[width=2.02cm, height=1.19cm]{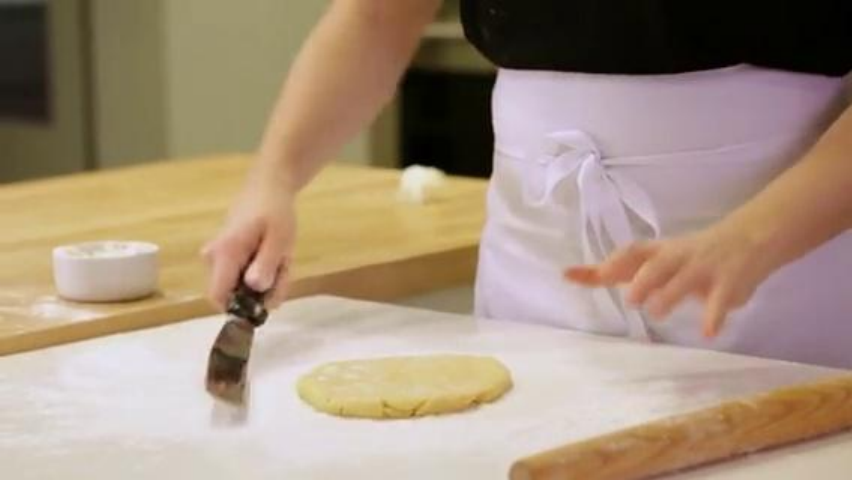} & 
                \hspace{-0.4cm}\includegraphics[width=2.02cm, height=1.19cm]{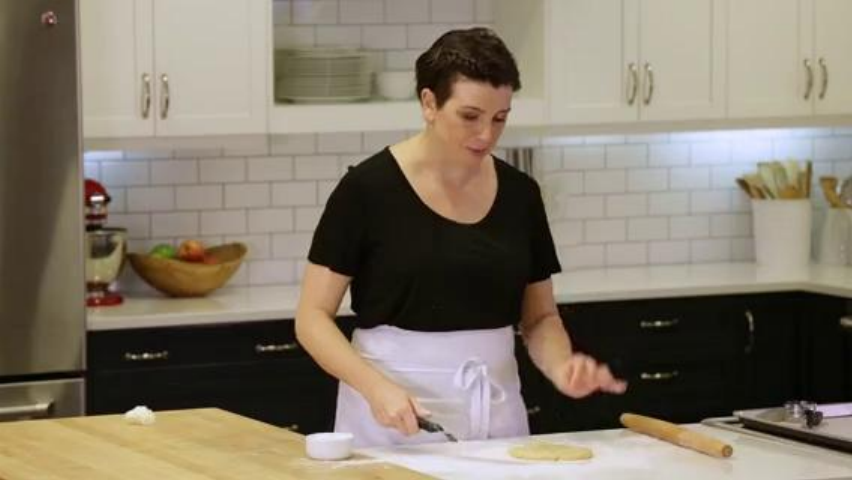}
            \end{tabular}};
        \end{tikzpicture}
        \vspace{-1.3cm}
        \begin{flushleft}
            \raisebox{0cm}{
                \begin{tabular}{ccccc}
                    \multicolumn{1}{c}{\hspace{-0.09cm}\parbox[c][0.5cm][c]{2cm}{\centering\scalebox{0.8}{\tiny\textcolor{red}{\text{}}}}}&
                    \multicolumn{1}{c}{\hspace{-0.3cm}\parbox[c][0.5cm][c]{2cm}{\centering\scalebox{0.8}{\tiny\text{}}}}&
                    \multicolumn{1}{c}{\hspace{-0.33cm}\parbox[c][0.5cm][c]{2cm}{\centering\scalebox{0.8}{\tiny\text{}}}}&
                    \multicolumn{1}{c}{\hspace{-0.4cm}\parbox[c][0.5cm][c]{2cm}{\centering\scalebox{0.8}{\tiny\text{}}}}
                \end{tabular}}
        \end{flushleft}
    \vspace{-0.55cm}
    \end{subtable}
    \begin{subtable}[t]{\linewidth} 
        \centering
        \begin{tikzpicture}
            \node (img) {\begin{tabular}{ccccc}
                
                \hspace{-0.59cm}\raisebox{-0.16cm}{
                \rotatebox{90}{\parbox[c][0.5cm][c]{1.5cm}{\centering\scalebox{0.8}{\small\text{Making pizza}}}}}&
                 \hspace{-0.49cm}\includegraphics[width=2.02cm, height=1.19cm]{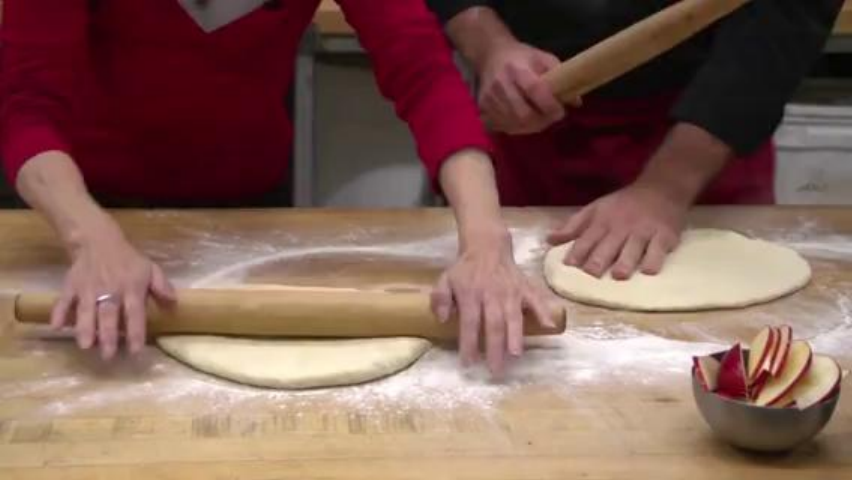} & 
                \hspace{-0.4cm}\includegraphics[width=2.02cm, height=1.19cm]{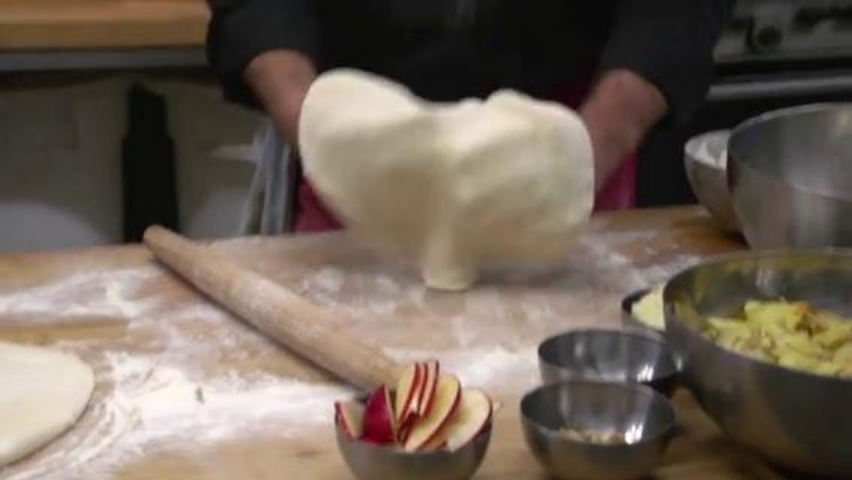} & 
                \hspace{-0.4cm}\includegraphics[width=2.02cm, height=1.19cm]{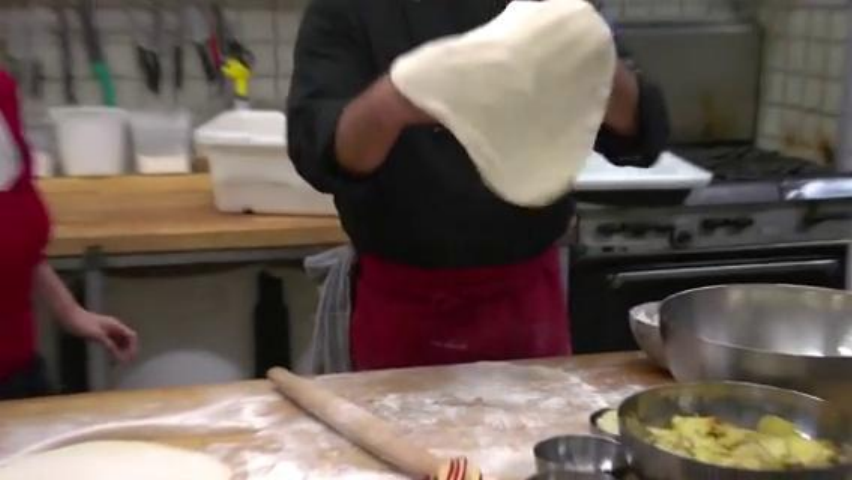} & 
                \hspace{-0.4cm}\includegraphics[width=2.02cm, height=1.19cm]{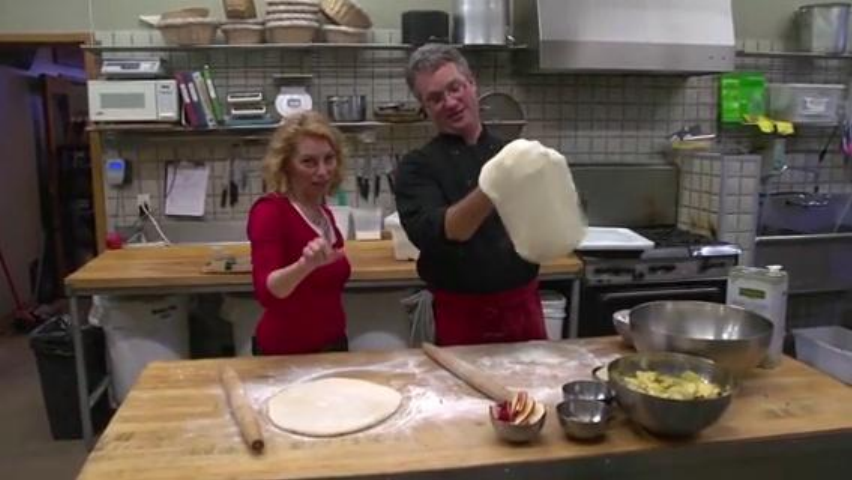}
            \end{tabular}};
        \end{tikzpicture}
        \vspace{-1.6cm}
        \begin{flushleft}
            \raisebox{0cm}{
                \begin{tabular}{cccc}
                    \multicolumn{1}{c}{\hspace{-0.09cm}\parbox[c][0.5cm][c]{2cm}{\centering\scalebox{0.8}{\tiny\textcolor{red}{\text{}}}}}&
                    \multicolumn{1}{c}{\hspace{-0.3cm}\parbox[c][0.5cm][c]{2cm}{\centering\scalebox{0.8}{\tiny\text{}}}}&
                    \multicolumn{1}{c}{\hspace{-0.33cm}\parbox[c][0.5cm][c]{2cm}{\centering\scalebox{0.8}{\tiny\text{}}}}&
                    \multicolumn{1}{c}{\hspace{-0.4cm}\parbox[c][0.5cm][c]{2cm}{\centering\scalebox{0.8}{\tiny\text{}}}}
                \end{tabular}}
        \end{flushleft}
        \vspace{-0.8cm}
        \begin{center}
            \raisebox{0cm}{
                \begin{tabular}{ccccc}
                    \multicolumn{5}{c}{\hspace{2.5cm}\parbox[c][0.5cm][c]{8cm}{\small(a) Ambiguous actions}}
                \end{tabular}}
        \end{center}
    \end{subtable}
    \vspace{-1.cm}
    \begin{subtable}[t]{\linewidth} 
        \centering        
        \vspace{-0.6cm}
        \begin{tikzpicture}
            \node (img) {\begin{tabular}{ccccccccccc}
                \hspace{-0.61cm}\raisebox{-0.15cm}{
                \rotatebox{90}{\parbox[c][0.5cm][c]{1.5cm}{\centering\scalebox{0.8}{\small\text{Swing legs}}}}}&
                \hspace{-0.45cm}\includegraphics[width=2.02cm, height=1.19cm]{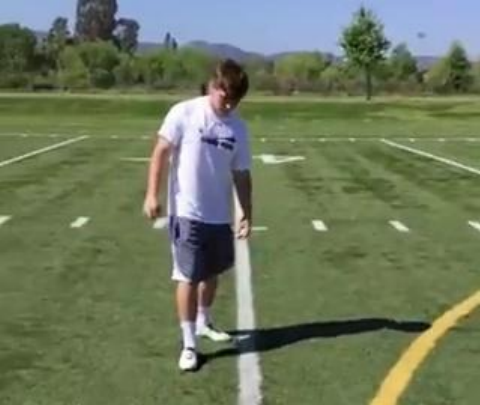}&
                \hspace{-2.135cm}\includegraphics[width=2.02cm, height=1.19cm]{pictures/fig1folder/swinglegs/frame_0017_xiugai.pdf}&
                \hspace{-2.135cm}\includegraphics[width=2.02cm, height=1.19cm]{pictures/fig1folder/swinglegs/frame_0017_xiugai.pdf}
                 & 
                \hspace{-0.34cm}\includegraphics[width=2.02cm, height=1.19cm]{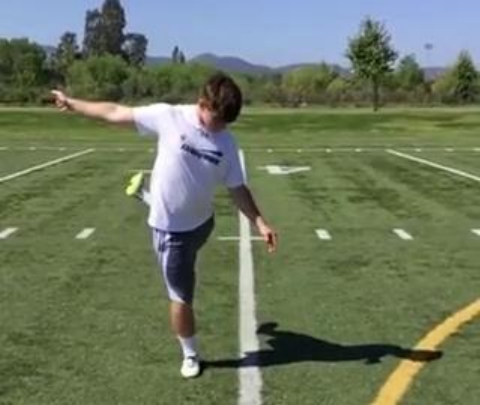}
                & 
                \hspace{-2.05cm}\includegraphics[width=2.02cm, height=1.19cm]{pictures/fig1folder/swinglegs/frame_0268_xiugai.pdf}
                & 
                \hspace{-2.05cm}\includegraphics[width=2.02cm, height=1.19cm]{pictures/fig1folder/swinglegs/frame_0268_xiugai.pdf}
                & 
                \hspace{-2.05cm}\includegraphics[width=2.02cm, height=1.19cm]{pictures/fig1folder/swinglegs/frame_0268_xiugai.pdf}
                & 
                \hspace{-2.05cm}\includegraphics[width=2.02cm, height=1.19cm]{pictures/fig1folder/swinglegs/frame_0268_xiugai.pdf}
                & 
                \hspace{-2.05cm}\includegraphics[width=2.02cm, height=1.19cm]{pictures/fig1folder/swinglegs/frame_0268_xiugai.pdf}
                 & 
                \hspace{-0.35cm}\includegraphics[width=1.37cm, height=1.19cm]{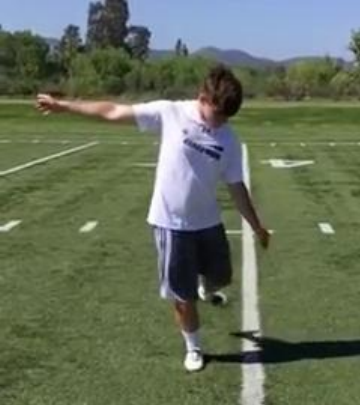}
            \end{tabular}};
        \end{tikzpicture}

        \vspace{-0.5cm}
        \begin{tikzpicture}
            \hspace{0.02cm}
            \node at (0cm,0) { 
                \begin{tikzpicture}
                    \draw[->] (0,0) -- (8.25,0) node[above left, xshift=0.3cm, yshift=-0.33cm] {\scalebox{0.8}{\small\text{time}}};
                    
                    \draw (0,-0.1) -- (0,0.1) node[below] {\scalebox{0.8}{\small\text{}}};
                    \draw (2.7,-0.1) -- (2.7,0.1) node[below] {\scalebox{0.8}{\small\text{}}};
                    \draw (6.75,-0.1) -- (6.75,0.1) node[below] {\scalebox{0.8}{\small\text{}}};
                \end{tikzpicture}
            };
        \end{tikzpicture}
        \vspace{-1.49cm}
        \begin{flushleft}
            \raisebox{0cm}{
                \begin{tabular}{ccc}
                    \multicolumn{1}{c}{\hspace{0.4cm}\parbox[c][0.5cm][c]{2cm}{\centering\scalebox{0.8}{\small\text{34\%}}}}&
                    \multicolumn{1}{c}{\hspace{1.1cm}\parbox[c][0.5cm][c]{2cm}{\centering\scalebox{0.8}{\small\text{52\%}}}}&
                    \multicolumn{1}{c}{\hspace{0.25cm}\parbox[c][0.5cm][c]{2cm}{\centering\scalebox{0.8}{\small\text{14\%}}}}
                \end{tabular}}
        \end{flushleft}

        \vspace{-0.61cm}
        \begin{flushleft}
            \raisebox{0cm}{
                \begin{tabular}{cccc}
                    \multicolumn{1}{c}{\hspace{0.15cm}\parbox[c][0.5cm][c]{2cm}{\centering\scalebox{0.8}{\small\text{Irrelevant frames}}}}&
                    &
                    \multicolumn{1}{c}{\hspace{-0.4cm}\parbox[c][0.5cm][c]{2cm}{\centering\scalebox{0.8}{\small\text{Swinging legs backward}}}}&
                    \multicolumn{1}{c}{\hspace{0.5cm}\parbox[c][0.5cm][c]{2cm}{\centering\scalebox{0.8}{\small\text{Swinging legs forward}}}}
                \end{tabular}}
        \end{flushleft}
    \vspace{0.75cm}
    \end{subtable}
    \vspace{0.5cm}
    \begin{subtable}[t]{\linewidth} 
        \vspace{-0.3cm}
        \centering
        \begin{tikzpicture}
            \node (img) {\begin{tabular}{ccccccccc}
                \hspace{-1.15cm}\raisebox{-0.16cm}{
                \rotatebox{90}{\parbox[c][0.5cm][c]{1.5cm}{\centering\scalebox{0.8}{\small\text{Long jump}}}}}&
                \hspace{-0.46cm}\includegraphics[width=2.02cm, height=1.19cm]{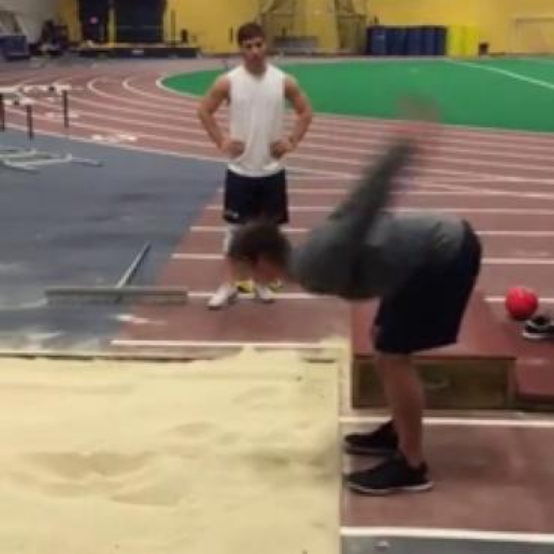} & 
                \hspace{-2cm}\includegraphics[width=2.02cm, height=1.19cm]{pictures/fig1folder/long/long22_xiugai.pdf} & 
                \hspace{-2cm}\includegraphics[width=2.02cm, height=1.19cm]{pictures/fig1folder/long/long22_xiugai.pdf} & \hspace{-2cm}\includegraphics[width=2.02cm, height=1.19cm]{pictures/fig1folder/long/long22_xiugai.pdf} & \hspace{-0.355cm}\includegraphics[width=1.75cm, height=1.19cm]{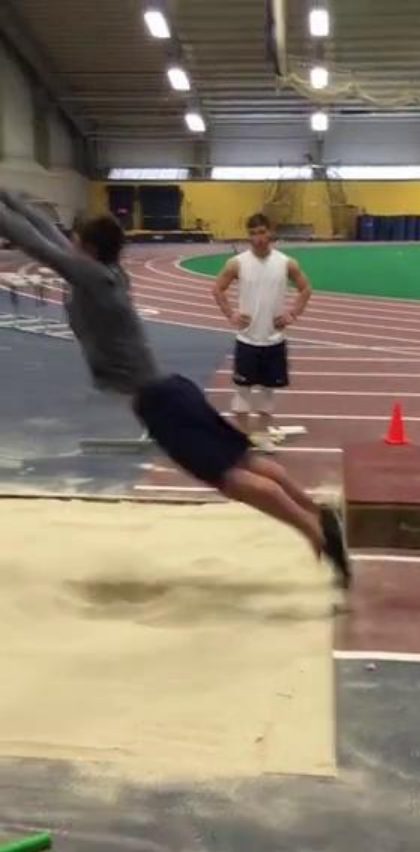} & \hspace{0.53cm}\includegraphics[width=2.02cm, height=1.19cm]{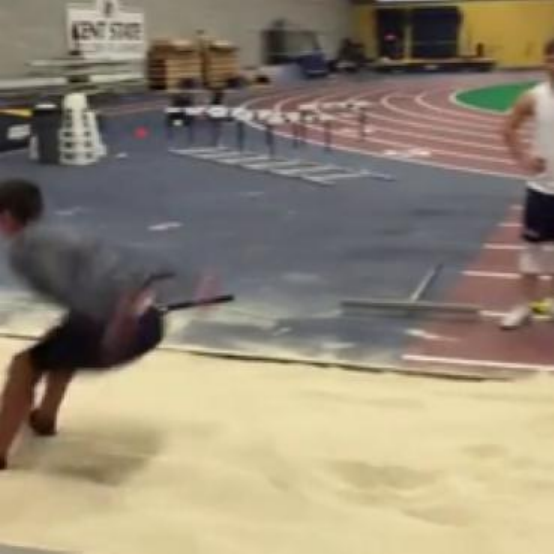}  
                & \hspace{-2.89cm}\includegraphics[width=2.02cm, height=1.19cm]{pictures/fig1folder/long/long62_xiugai.pdf}  
                & \hspace{-2.89cm}\includegraphics[width=2.02cm, height=1.19cm]{pictures/fig1folder/long/long62_xiugai.pdf} 
            \end{tabular}};
        \end{tikzpicture}
        
        \vspace{-0.39cm}
        \begin{tikzpicture}
            \hspace{0.14cm}
            \draw[->] (0,0) -- (8.25,0) node[above left, xshift=0.3cm, yshift=-0.33cm] {\scalebox{0.8}{\small\text{time}}};
            
            \draw (0,-0.1) -- (0,0.1) node[below] {\scalebox{0.8}{\small\text{}}};
            \draw (3.4,-0.1) -- (3.4,0.1) node[below] {\scalebox{0.8}{\small\text{}}};
            \draw (5.22,-0.1) -- (5.22,0.1) node[below] {\scalebox{0.8}{\small\text{}}};
        \end{tikzpicture}
        \vspace{-0.99cm}
        \begin{flushleft}
            \raisebox{0cm}{
                \begin{tabular}{ccc}
                    \multicolumn{1}{c}{\hspace{0.8cm}\parbox[c][0.5cm][c]{2cm}{\centering\scalebox{0.8}{\small\text{43\%}}}}&
                    \multicolumn{1}{c}{\hspace{0.15cm}\parbox[c][0.5cm][c]{2cm}{\centering\scalebox{0.8}{\small\text{22\%}}}}&
                    \multicolumn{1}{c}{\hspace{0.cm}\parbox[c][0.5cm][c]{2cm}{\centering\scalebox{0.8}{\small\text{35\%}}}}
                \end{tabular}}
        \end{flushleft}

        \vspace{-0.61cm}
        \begin{flushleft}
            \raisebox{0cm}{
                \begin{tabular}{ccc}
                    \multicolumn{1}{c}{\hspace{0.65cm}\parbox[c][0.5cm][c]{2cm}{\centering\scalebox{0.8}{\small\text{Crouch, swing arms}}}}&
                    \multicolumn{1}{c}{\hspace{0.22cm}\parbox[c][0.5cm][c]{2cm}{\centering\scalebox{0.8}{\small\text{Extend legs}}}}&
                    \multicolumn{1}{c}{\hspace{-0.1cm}\parbox[c][0.5cm][c]{2cm}{\centering\scalebox{0.8}{\small\text{Land with bent knees}}}}
                \end{tabular}}
        \end{flushleft}
        \vspace{-0.75cm}
        \begin{center}
            \raisebox{0cm}{
                \begin{tabular}{ccccc}
                    \multicolumn{5}{c}{\hspace{2.5cm}\parbox[c][0.5cm][c]{8cm}{\small(b) Non-uniform actions}}
                \end{tabular}}
        \end{center}
    \end{subtable}
    \vspace{-0.8cm}
    \caption{\it \small 
    Example videos with ambiguous or non-uniform actions. Each row displays sample frames of a video with its class name in the left column. (a) Ambiguous actions. Both `Baking cookies' and `Making pizza' are actions performed in the kitchen which has a similar visual appearance, and they share atomic actions, \eg, the rolling dough. (b) Non-uniform actions. Multiple atomic actions distribute unevenly within a video to form an action. Moreover, an atomic action can be non-directly related to the global video semantics, \eg, `Stand' in the video of `Swing legs'. The axis represents the percentage of frames associated with atomic actions.
    }
    \vspace{-4mm}
    \label{fig:overview}
\end{table}

\begin{table*}[t]
		\centering
        \captionsetup{type=figure}
\begin{minipage}{\textwidth}
    \vspace{-0.8cm}
    \hspace{10cm} 
    \begin{minipage}[t]{0.25\textwidth}
        \vspace{0.5cm} 
        \hspace{0.2cm}
        \begin{tikzpicture}[baseline]
            \begin{axis}[
                grid,
                ybar,
                xtick=\empty,
                ymin=69,
                ymax=100,
                scaled y ticks=false,
                title style={at={(0,1)}, anchor=north west, yshift=-1.5ex},
                ylabel={\scalebox{0.8}{\small Top-1 Acc (\%)}},
                bar width=4pt,
                width=0.9\textwidth,
                height=0.6\textwidth,
                enlargelimits=0.15,
                y tick label style={font=\tiny, xshift=0cm},
                x tick label style={font=\tiny, yshift=0cm},
                label style={font=\tiny},
                y label style={at={(axis description cs:0.30,.5)},anchor=south, font=\tiny},
                x label style={at={(axis description cs:0.5,0.1)},anchor=north, font=\tiny},
                scaled y ticks=false, 
                /pgf/number format/fixed, 
                xtick={0,1,2,3,4,5,6,7,8, 9}, 
                xtick style={/pgfplots/major tick length=0pt}, 
                xticklabels={,,,,,,,,,}, 
                legend style={
                    legend cell align=left,
                    legend pos=outer north east,
                    legend columns=1
                }
            ]
            \addplot [
                forestgreen, 
                fill=forestgreen!30!white
            ]table [x expr=\coordindex, y index=0, col sep=tab] {average_accuracies_bike_top1_top10.txt};
            \end{axis}
        \end{tikzpicture}
    \end{minipage}%
    \begin{minipage}[t]{0.25\textwidth}
        \vspace{0.50cm} 
        \hspace{-0.5cm}
        \begin{tikzpicture}[baseline]
            \begin{axis}[
                grid,
                ybar,
                xtick=\empty,
                ymin=80,
                ymax=87,
                title style={at={(0,1)}, anchor=north west, yshift=-1.5ex},
                ylabel={\scalebox{0.8}{\small Top-1 Acc (\%)}},
                bar width=4pt,
                width=0.9\textwidth,
                height=0.6\textwidth,
                enlargelimits=0.15,
                y tick label style={font=\tiny, xshift=0cm},
                x tick label style={font=\tiny, yshift=0cm},
                label style={font=\tiny},
                y label style={at={(axis description cs:0.30,.5)},anchor=south, font=\tiny},
                x label style={at={(axis description cs:0.5,0.1)},anchor=north, font=\tiny},
                scaled y ticks=false, 
                /pgf/number format/fixed, 
                legend pos=outer north east, 
                xtick={0,1,2,3,4,5,6,7,8, 9}, 
                xtick style={/pgfplots/major tick length=0pt}, 
                xticklabels={,,,,,,,,,}, 
                legend style={
                    legend cell align=left,
                    legend pos=outer north east,
                    legend columns=1
                }
            ]
             \addplot [
                forestgreen, 
                fill=forestgreen!30!white
            ] coordinates {
                (0, 86.45) 
                (1, 84.92)
                (2, 84.62)
                (3, 85.35)
                (4, 83.94)
                (5, 82.23)
                (6, 80.77)
                (7, 81.93)
                (8, 80.34)
                (9, 80.09)
            };
            \end{axis}
        \end{tikzpicture}
    \end{minipage}%
    \vspace{-0.3cm}
    
    \hspace{10cm}
    \begin{minipage}[t]{0.25\textwidth}
        \vspace{-0.1cm} 
        \hspace{0.25cm}
        \begin{tikzpicture}[baseline]
            \begin{axis}[
                grid,
                ybar,
                xtick=\empty,
                ymin=0,
                ymax=3.5,
                scaled y ticks=false,
                title style={at={(0,1)}, anchor=north west, yshift=-1.5ex},
                ylabel={\scalebox{0.8}{\small Increase}},
                bar width=4pt,
                width=0.9\textwidth,
                height=0.6\textwidth,
                enlargelimits=0.15,
                y tick label style={font=\tiny, xshift=0cm},
                x tick label style={font=\tiny, yshift=0cm},
                label style={font=\tiny},
                y label style={at={(axis description cs:0.30,.5)},anchor=south, font=\tiny},
                x label style={at={(axis description cs:0.5,0.1)},anchor=north, font=\tiny},
                scaled y ticks=false, 
                /pgf/number format/fixed, 
                xtick={0,1,2,3,4,5,6,7,8, 9}, 
                xtick style={/pgfplots/major tick length=0pt}, 
                xticklabels={,,,,,,,,,}, 
                legend style={
                    legend cell align=left,
                    legend pos=outer north east,
                    legend columns=1
                }
            ]
            \addplot [
                blue, fill=blue!30!white
            ] coordinates {
                (0, 1.10) 
                (1, 1.15)
                (2, 1.46)
                (3, 2.47)
                (4, 1.66)
                (5, 2.08)
                (6, 1.15)
                (7, 2.09)
                (8, 2.24)
                (9, 1.47)
            };
            \end{axis}
        \end{tikzpicture}
    \end{minipage}%
    \begin{minipage}[t]{0.25\textwidth}
        \vspace{0.05cm} 
        \hspace{-0.45cm}
        \begin{tikzpicture}[baseline]
            \begin{axis}[
                grid,
                ybar,
                xtick=\empty,
                ymin=0,
                ymax=6.5,
                title style={at={(0,1)}, anchor=north west, yshift=-1.5ex},
                ylabel={\scalebox{0.8}{\small Increase }},
                bar width=4pt,
                width=0.9\textwidth,
                height=0.6\textwidth,
                enlargelimits=0.15,
                y tick label style={font=\tiny, xshift=0cm},
                x tick label style={font=\tiny, yshift=0cm},
                label style={font=\tiny},
                y label style={at={(axis description cs:0.30,.5)},anchor=south, font=\tiny},
                x label style={at={(axis description cs:0.5,0.1)},anchor=north, font=\tiny},
                scaled y ticks=false, 
                /pgf/number format/fixed, 
                legend pos=outer north east, 
                xtick={0,1,2,3,4,5,6,7,8, 9}, 
                xtick style={/pgfplots/major tick length=0pt}, 
                xticklabels={,,,,,,,,,}, 
                legend style={
                    legend cell align=left,
                    legend pos=outer north east,
                    legend columns=1
                }
            ]
           \addplot [
                blue, fill=blue!30!white
            ] coordinates {
                (0, 4.58) 
                (1, 3.79)
                (2, 3.17)
                (3, 5.31)
                (4, 2.81)
                (5, 2.81)
                (6, 2.81)
                (7, 2.81)
                (8, 2.81)
                (9, 2.80)
            };
            \end{axis}
        \end{tikzpicture}
    \end{minipage}%
    \vspace{-0.4cm}
    
    \hspace{10cm}
    \vspace{-0.4cm}
    \begin{minipage}[t]{0.25\textwidth}
        \vspace{0cm} 
        \hspace{0.25cm}
        \begin{tikzpicture}[baseline]
            \begin{axis}[
                grid,
                ybar,
                xtick=\empty,
                ymin=0,
                ymax=3.5,
                scaled y ticks=false,
                title style={at={(0,1)}, anchor=north west, yshift=-1.5ex},
                ylabel={\scalebox{0.8}{\small Increase }},
                bar width=4pt,
                width=0.9\textwidth,
                height=0.6\textwidth,
                enlargelimits=0.15,
                y tick label style={font=\tiny, xshift=0cm},
                x tick label style={font=\tiny, yshift=0cm},
                label style={font=\tiny},
                y label style={at={(axis description cs:0.30,.5)},anchor=south, font=\tiny},
                x label style={at={(axis description cs:0.5,0.1)},anchor=north, font=\tiny},
                scaled y ticks=false, 
                /pgf/number format/fixed, 
                xtick={0,1,2,3,4,5,6,7,8, 9}, 
                xtick style={/pgfplots/major tick length=0pt}, 
                xticklabels={,,,,,,,,,}, 
                legend style={
                    legend cell align=left,
                    legend pos=outer north east,
                    legend columns=1
                }
            ]
            \addplot [
                red, 
                fill=red!30!white
            ]coordinates {
                (0, 1.22) 
                (1, 0.79)
                (2, 1.16)
                (3, 2.40)
                (4, 1.60)
                (5, 1.95)
                (6, 1.09)
                (7, 1.78)
                (8, 1.93)
                (9, 2.51)
            };
            \end{axis}
        \end{tikzpicture}
    \end{minipage}%
    \begin{minipage}[t]{0.25\textwidth}
        \vspace{0.15cm} 
        \hspace{-0.45cm}
        \begin{tikzpicture}[baseline]
            \begin{axis}[
                grid,
                ybar,
                xtick=\empty,
                ymin=0,
                ymax=6.5,
                title style={at={(0,1)}, anchor=north west, yshift=-1.5ex},
                ylabel={\scalebox{0.8}{\small Increase }},
                bar width=4pt,
                width=0.9\textwidth,
                height=0.6\textwidth,
                enlargelimits=0.15,
                y tick label style={font=\tiny, xshift=0cm},
                x tick label style={font=\tiny, yshift=0cm},
                label style={font=\tiny},
                y label style={at={(axis description cs:0.30,.5)},anchor=south, font=\tiny},
                x label style={at={(axis description cs:0.5,0.1)},anchor=north, font=\tiny},
                scaled y ticks=false, 
                /pgf/number format/fixed, 
                legend pos=outer north east, 
                xtick={0,1,2,3,4,5,6,7,8, 9}, 
                xtick style={/pgfplots/major tick length=0pt}, 
                xticklabels={,,,,,,,,,}, 
                legend style={
                    legend cell align=left,
                    legend pos=outer north east,
                    legend columns=1
                }
            ]
           \addplot [
                red, 
                fill=red!30!white
            ]coordinates {
                (0, 3.54) 
                (1, 2.87)
                (2, 1.95)
                (3, 1.83)
                (4, 2.44)
                (5, 1.71)
                (6, 2.08)
                (7, 2.08)
                (8, 2.08)
                (9, 2.07)
            };
            \end{axis}
        \end{tikzpicture}
    \end{minipage}%
    \vspace{0.1cm}
    
    \hspace{10cm}
    \begin{minipage}[t]{0.25\textwidth}
        \vspace{-0.1cm} 
        \hspace{0.25cm}
        \begin{tikzpicture}[baseline]
            \begin{axis}[
                grid = major,
                ybar,
                xtick=\empty,
                xlabel={\scalebox{0.8}{\small Ambiguous Score}},
                ymin=0,
                ymax=3.5,
                title style={at={(0,1)}, anchor=north west, yshift=-1.5ex},
                ylabel={\scalebox{0.8}{\small Increase }},
                bar width=4pt,
                width=0.9\textwidth,
                height=0.6\textwidth,
                enlargelimits=0.15,
                y tick label style={font=\tiny, xshift=0cm},
                x tick label style={font=\tiny, yshift=0.1cm},
                label style={font=\tiny},
                y label style={at={(axis description cs:0.30,.5)},anchor=south, font=\tiny},
                x label style={at={(axis description cs:0.5,0.2)},anchor=north, font=\tiny},
                scaled y ticks=false, 
                /pgf/number format/fixed, 
                legend pos=outer north east, 
                xtick={0,1,2,3,4,5,6,7,8, 9}, 
                xtick style={/pgfplots/major tick length=0pt}, 
                xticklabels={1,,,,5,,,,,10} 
            ]
            \addplot [
                orange, 
                fill=orange!30!white
            ] coordinates {
                (0, 1.77) 
                (1, 1.33)
                (2, 1.95)
                (3, 3.21)
                (4, 1.91)
                (5, 2.87)
                (6, 1.69)
                (7, 2.46)
                (8, 3.44)
                (9, 2.81)
            };
            \end{axis}
        \end{tikzpicture}
    \end{minipage}%
    \begin{minipage}[t]{0.25\textwidth}
        \vspace{0.05cm} 
        \hspace{-0.45cm}
        \begin{tikzpicture}[baseline]
            \begin{axis}[
                grid,
                ybar,
                xtick=\empty,
                xlabel={\scalebox{0.8}{\small Non-uniform Score}},
                ymin=0,
                ymax=6.5,
                title style={at={(0,1)}, anchor=north west, yshift=-1.5ex},
                ylabel={\scalebox{0.8}{\small Increase }},
                bar width=4pt,
                width=0.9\textwidth,
                height=0.6\textwidth,
                enlargelimits=0.15,
                enlarge x limits=0.15,
                y tick label style={font=\tiny, xshift=0cm},
                x tick label style={font=\tiny, yshift=0.1cm},
                label style={font=\tiny},
                y label style={at={(axis description cs:0.30,.5)},anchor=south, font=\tiny},
                x label style={at={(axis description cs:0.5,0.2)},anchor=north, font=\tiny},
                scaled y ticks=false, 
                /pgf/number format/fixed, 
                legend pos=outer north east, 
                xtick={0,1,2,3,4,5,6,7,8, 9}, 
                xticklabels={1,,,,5,,,,,10}, 
                xtick style={/pgfplots/major tick length=0pt}, 
            ]
           \addplot [
                orange, 
                fill=orange!30!white
            ] coordinates {
                (0, 6.35) 
                (1, 4.76)
                (2, 5.07)
                (3, 3.48)
                (4, 3.79)
                (5, 2.87)
                (6, 2.87)
                (7, 2.87)
                (8, 2.87)
                (9, 2.86)
            };
            \end{axis}
            \end{tikzpicture}
        \end{minipage}
    \end{minipage}

    \begin{minipage}[t]{\textwidth}
        \vspace{-5.8cm}
        \begin{minipage}[t]{0.75\textwidth}
            \begin{tabular}{cccccc}
            \hspace{0.20cm}\raisebox{-0.1cm}{\rotatebox{90}{\parbox[c]{1.cm}{\centering \scalebox{0.8}{\small\text{Making a sandwich}}}}} &
            \hspace{-0.35cm}\includegraphics[width =1.8cm, height=1.8cm]{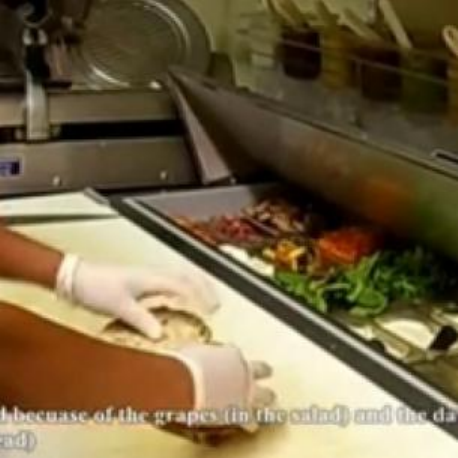} & \hspace{-0.3cm}\includegraphics[width =1.8cm,height=1.8cm]{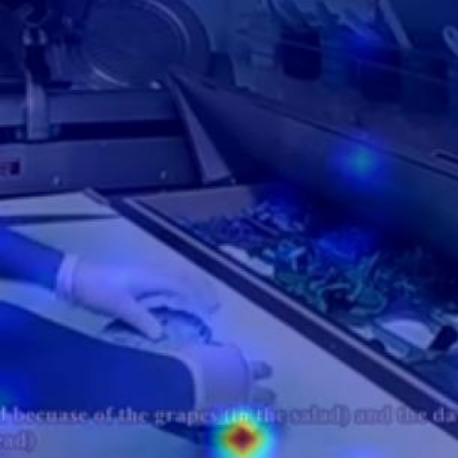} & \hspace{-0.3cm}\includegraphics[width =1.8cm,height=1.8cm]{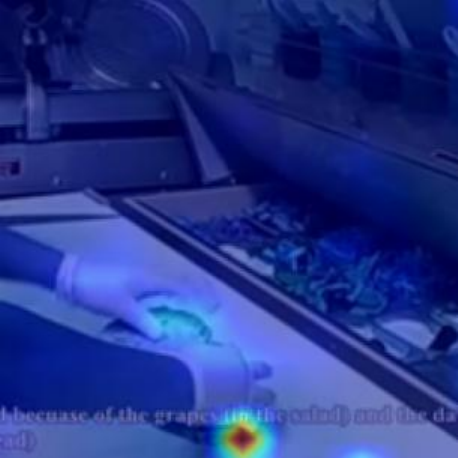} & 
            \hspace{-0.3cm}\includegraphics[width =1.8cm,height=1.8cm]{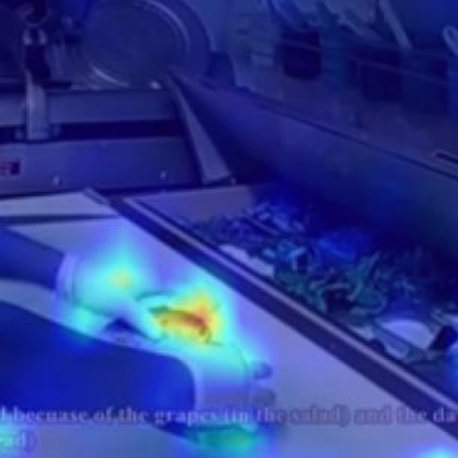} &
            \hspace{-0.3cm}\includegraphics[width =1.8cm,height=1.8cm]{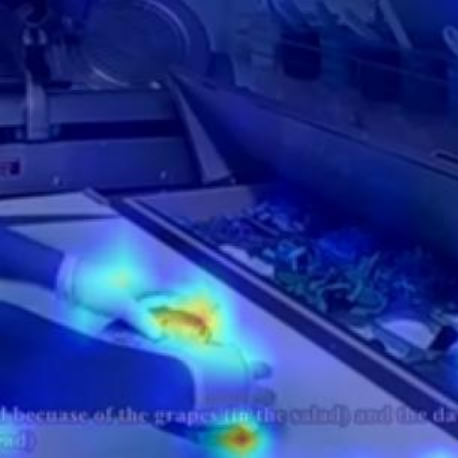}
            \\
            \hspace{0.20cm}\raisebox{0.4cm}{\rotatebox{90}{\parbox[c]{1.0cm}{\centering \scalebox{0.8}{\small\text{Pull ups}}}}} &
            \hspace{-0.35cm}\includegraphics[width =1.8cm, height=1.8cm]{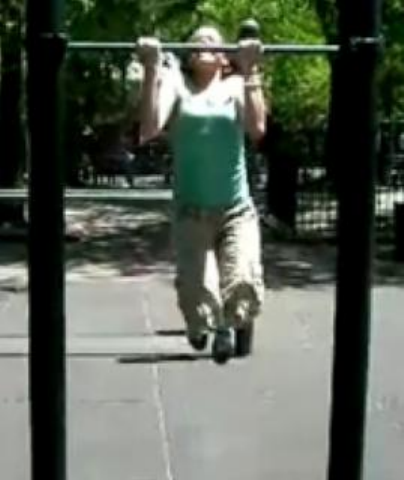} & \hspace{-0.3cm}\includegraphics[width =1.8cm,height=1.8cm]{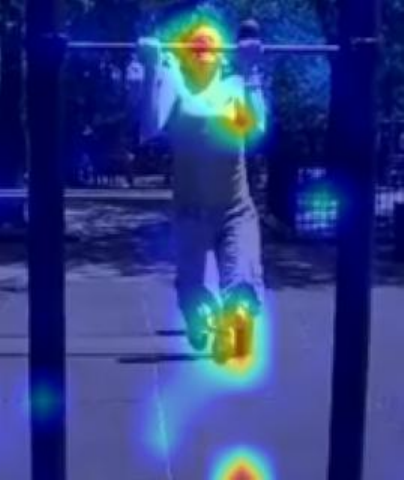} & \hspace{-0.3cm}\includegraphics[width =1.8cm,height=1.8cm]{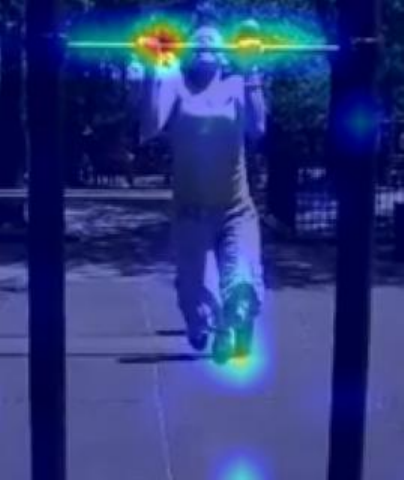} &
            \hspace{-0.3cm}\includegraphics[width =1.8cm,height=1.8cm]{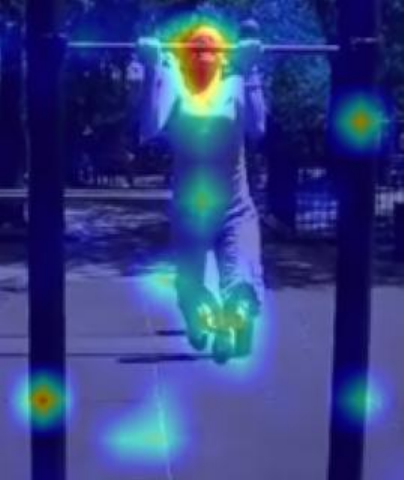} &
            \hspace{-0.3cm}\includegraphics[width =1.8cm,height=1.8cm]{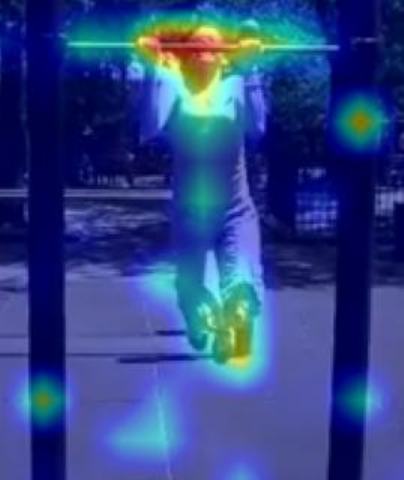} 
            \\
            \hspace{0.20cm}\raisebox{0.1cm}{\rotatebox{90}{\parbox[c]{1.0cm}{\centering \scalebox{0.8}{\small\text{Shearing sheep}}}}} &
            \hspace{-0.35cm}\includegraphics[width =1.8cm, height=1.8cm]{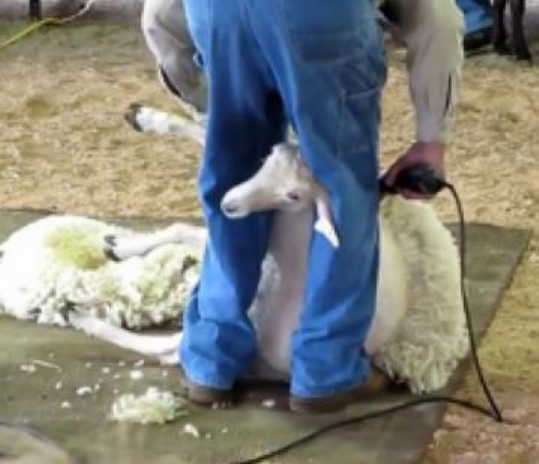} & \hspace{-0.3cm}\includegraphics[width =1.8cm,height=1.8cm]{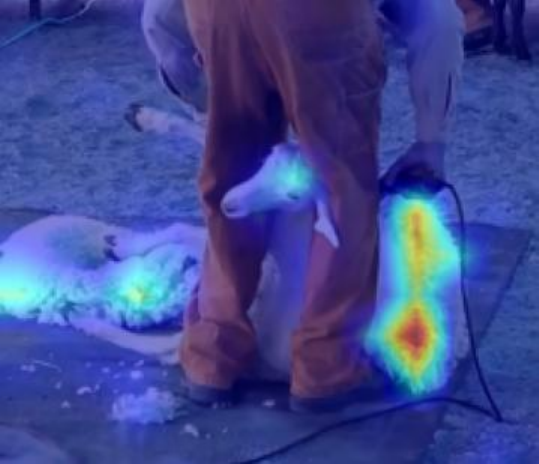} & \hspace{-0.3cm}\includegraphics[width =1.8cm,height=1.8cm]{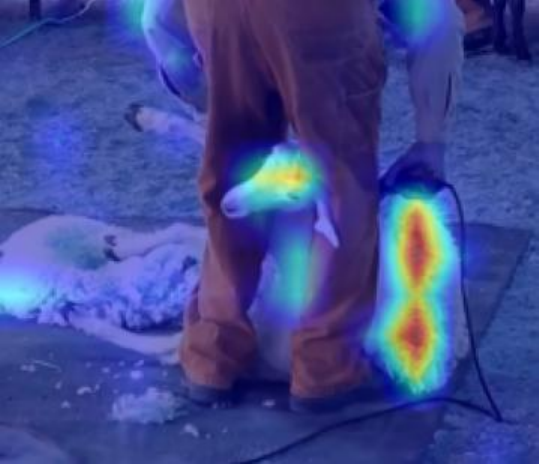} &
            \hspace{-0.3cm}\includegraphics[width =1.8cm,height=1.8cm]{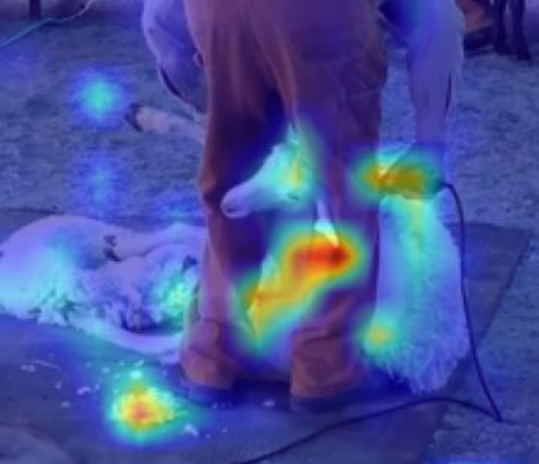} &
            \hspace{-0.3cm}\includegraphics[width =1.8cm,height=1.8cm]{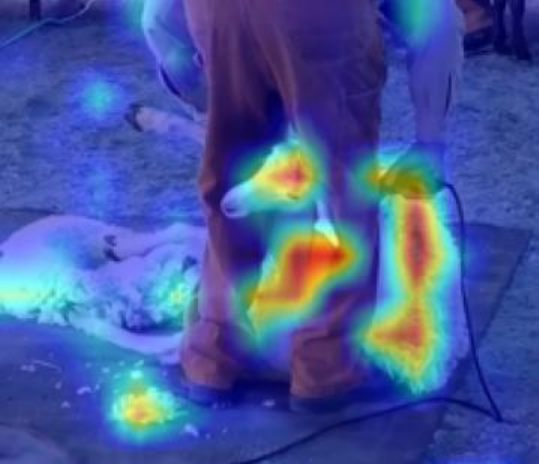}
            \end{tabular}
        \end{minipage}
    \end{minipage}
    \begin{minipage}[t]{\textwidth}
        \vspace{-0.55cm}
        \hspace{0.8cm}
        \begin{minipage}[t]{0.65\textwidth}
            \raisebox{0cm}{
                \begin{tabular}{cccccc}
                    \hspace{-0.1cm}\scalebox{0.8}{\small (a) Frame} & 
                    \hspace{0.5cm}\scalebox{0.8}{\small (b) Single} & 
                    \hspace{0.5cm}\scalebox{0.8}{\small (c) Coarse} & 
                    \hspace{0.60cm}\scalebox{0.8}{\small (d) Fine} & 
                    \hspace{0.23cm}\scalebox{0.8}{\small (e) Coarse+Fine} &
                    \hspace{3.13cm}\scalebox{0.8}{\small (f) Performance}
                \end{tabular}
            }
        \end{minipage}
    \end{minipage}
    \vspace{-0.7cm}
    \caption{\it \small Examples that illustrate the ineffectiveness of directly aligning video frames and global text, from (a-e) heatmaps and (f) classification performance improvements. (a) Video frames with their class name annotated at the left. (b) Heatmaps of the \textcolor{forestgreen}{single granularity-based (single) method} \cite{wu2023bidirectional}, which are limited by the granularity difference between the video frame and global text. (c-e) Heatmaps using \textcolor{blue}{coarse video embedding}, \textcolor{red}{fine-grained video embedding}, and \textcolor{orange}{the fusion of the two}. (f) We show the performance distributions (Top-1 Accuracy) over ambiguous scores and non-uniform scores of videos for (b-e) on the whole set. The first row is the absolute performance for \textcolor{forestgreen}{(b)}, and the last three rows are improvements of  \textcolor{blue}{(c)}, \textcolor{red}{(d)}, and \textcolor{orange}{(e)} concerning \textcolor{forestgreen}{(b)}. Refer to our supplementary materials for details on the computation of the ambiguous score and the non-uniform score.
    }
    \label{fig:heatmap}
    \vspace{-5mm}
\end{table*}	

The field of fine-grained video action recognition has garnered increasing attention due to its broad applicability in areas such as sports analytics~\cite{xue2023clip}, human-computer interaction~\cite{hassan2021populating}, surveillance~\cite{wu2023boosting}, and video understanding~\cite{li2024mvbench}. In contrast to standard action recognition, fine-grained video action recognition necessitates a more detailed understanding of actions with similar appearances, demanding greater precision in capturing key frames. 

The advent of large language models has revealed that a robust multimodal encoder like CLIP~\cite{clip} can consolidate significantly more potent learned embeddings compared to manually crafted embeddings for action recognition~\cite{wang2021actionclip, lin2022frozen, qing2023disentangling, arnab2021vivit, videoswin_22_cvpr, pan2022st, jeong2023winclip}. These methods involve using CLIP's visual encoder to extract video embeddings and its textual encoder to extract text embeddings. This can be done using learnable text prompts~\cite{ju2022prompting} or class-formatted text prompts derived from video class names~\cite{feng2023text, wu2023revisiting}. The video and text embeddings are then aligned for video classification. However, the potential of the textual encoder has not been fully explored.


In order to effectively integrate textual information, recent works~\cite{wu2023bidirectional, ni2022expanding, chen2023FDT} have proposed the identification of video frames that closely correspond to the class-formatted text prompt for computing video embeddings. These methods make two assumptions: 1) each global video semantics should have identical atomic actions. 2) all atomic actions within a video should be closely related to the global video semantics. However, these assumptions are often violated, making the embedding inaccurate. For instance, in Fig.~\ref{fig:overview} (a), the videos `Baking cookies' and `Making pizza' feature overlapping atomic actions. In Fig.~\ref{fig:overview} (b), an atomic action `Stand' can be irrelevant to the global video semantics `Swing legs'. 

Moreover, the non-uniform distribution nature of atomic actions has been overlooked previously, \ie, the duration of each atomic action can vary, making the key frames corresponding to atomic actions non-uniformly distributed. Based on the above observation, using the global video semantics directly for aligning videos and text prompts without making any semantic distinctions can lead to misunderstandings due to their granularity discrepancy.




In this paper, we address these issues and propose a multi-granularity framework. The core of our framework involves decomposing `a global text that describes the global video semantics into fine-grained sub-texts' and `a video action into multiple atomic actions'. This allows for both coarse and fine-grained identification of key frames in videos, thereby enhancing video action recognition. 
 
Inspired by the concept of storyboarding, which breaks down a script into individual shots, we enhance the global text by generating detailed descriptions using a pre-trained large language model (LLM). These fine-grained descriptions, \ie, sub-texts, capture common atomic actions depicted in videos, requiring only their class names and utilizing our designed question prompts. 
To fit the flexibility of various pre-trained LLM and question prompt formats, we design a text prompt perplexity metric to measure the diversity among sub-texts and the similarity between the sub-text to the global text for filtering sub-texts. It provides an effective schema for selecting sub-texts to train the video action recognition model.

We then use global texts and sub-texts to coarsely and fine-grainedly compute a video embedding for videos with ambiguous or non-uniform actions. Specifically, we augment the global text with sub-texts in the embedding space of CLIP by abounding described video actions with sub-texts, which also decreases the granularity of the global text. This augmented global text is used to weight video frames in embedding space and compute a coarse video embedding. In Fig.~\ref{fig:heatmap} (c), we show the effectiveness of the augmented global text. Our augmented global text draws the network attention to more fully cover the areas with human action than the past method limited by the granularity difference (Fig.~\ref{fig:heatmap} (b)).

In videos with non-uniform atomic actions, we find video frames that perform atomic actions with the sub-texts for adaptively computing a fine-grained video embedding. 
Refer to Fig.~\ref{fig:heatmap} (d) for network attention with the sub-texts, finding heatmaps that better focus on regions with actions.
We then fuse the coarse and fine-grained video embedding to compute a video embedding that improves the classification performance on ambiguous and non-uniform videos. Heatmaps of our network are in Fig.~\ref{fig:heatmap} (e), and improvements over videos with different ambiguous and non-uniform scores are in Fig.~\ref{fig:heatmap} (f). 

We rigorously validate our approach across different scenarios, including supervised, few-shot, and zero-shot video action recognition. Our method delivers top-notch performance in all these scenarios, showcasing the effectiveness of our framework.

Our main contributions are:
\begin{itemize}
\itemsep-0.15em
    \item A multi-granularity framework for transferring CLIP trained on image-text pair to video action recognition.
    \item A schema for automatically decomposing a video action into common atomic actions to provide fine-grained knowledge of the video action to CLIP.
    \item A coarse and fine-grained video embedding module for videos with ambiguous and non-uniform actions.
\end{itemize}

\begin{figure*}[t]
\begin{center}
\includegraphics{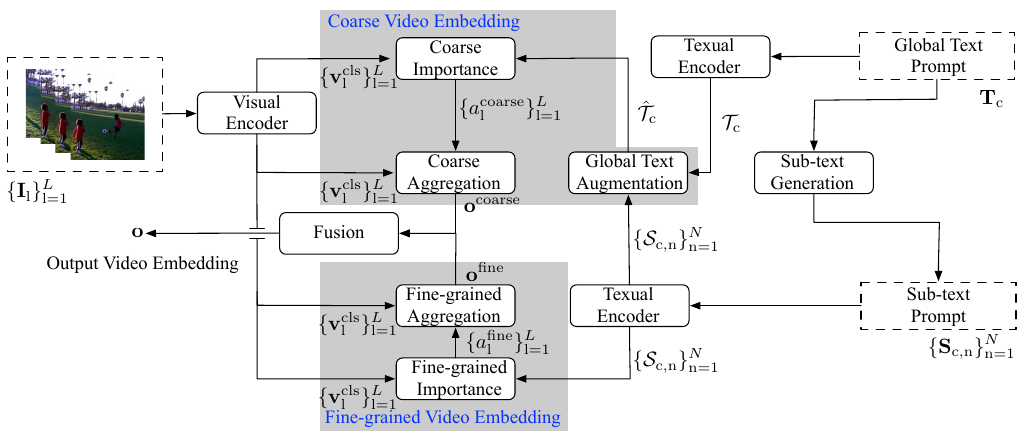}
\end{center}
\vspace{-0.47cm}
\caption{\it \small An overview of our framework for video action recognition. We extend CLIP for classifying the video \textnormal{$\{\mathbf{I}_{\text{l}}\}_{\text{l}=1}^{L}$} with \textnormal{$L$} frames by computing a video embedding from frame embedding   \textnormal{$\{\mathbf{v}_{\text{l}}^\text{cls}\}_{\text{l}=1}^{L}$} extracted with a visual encoder of CLIP in three steps. 
i) We decompose the global text prompt \textnormal{$\mathbf{T}_{\text{c}}$} that describes the class semantic of action \textnormal{$\text{c}$} into descriptions of atomic action (\ie, sub-text prompts) by using a pre-trained large language model. The global text prompt \textnormal{$\mathbf{T}_{\text{c}}$} and sub-text prompts are then embedded by the textual encoder of CLIP for extracting embeddings of \textnormal{$\mathcal{T}_{\text{c}}$} and \textnormal{$\{\mathcal{S}_{\text{c}, \text{n}}\}_{\text{l}=1}^{L}$}.
ii) A coarse video embedding is extracted by augmenting the global text embedding \textnormal{$\mathcal{T}_{\text{c}}$} with the sub-text embedding \textnormal{$\{\mathcal{S}_{\text{c}, \text{n}}\}_{\text{l}=1}^{L}$}, 
calculating coarse importance of the frame embedding \textnormal{$\{\mathbf{v}_{\text{l}}^\text{cls}\}_{\text{l}=1}^{L}$} with the augmented global text embedding \textnormal{$\hat{\mathcal{T}}_{\text{c}}$}, and using the importance \textnormal{$\{a_{\text{l}}\}_{\text{l}=1}^{L}$} to aggregate a coarse video embedding \textnormal{$\mathbf{o}^\text{coarse}$} from frame embedding \textnormal{$\{\mathbf{v}_{\text{l}}^\text{cls}\}_{\text{l}=1}^{L}$}. 
iii) Similar to the coarse video embedding, we get a fine-grained video embedding \textnormal{$\mathbf{o}^\text{fine}$} by calculating fine-grained importance \textnormal{$\{a_{\text{l}}^\text{fine}\}_{\text{l}=1}^{L}$}  of the frame embedding \textnormal{$\{\mathbf{v}_{\text{l}}^\text{cls}\}_{\text{l}=1}^{L}$} with sub-text embedding \textnormal{$\{\mathcal{S}_{\text{c}, \text{n}}\}_{\text{l}=1}^{L}$} for aggregating frame embedding \textnormal{$\{\mathbf{v}_{\text{l}}^\text{cls}\}_{\text{l}=1}^{L}$}.
The coarse video embedding \textnormal{$\mathbf{o}^\text{coarse}$} and fine-grained video embedding \textnormal{$\mathbf{o}^\text{fine}$} are fused to video embedding \textnormal{$\mathbf{o}$} for action recognition.
}
\label{fig:framework}
\vspace{-4mm}
\end{figure*}

\section{Related Works}
We review video action recognition methods that focus on visual and visual-text representation learning.

\vspace{+0.5mm}
\noindent{\bf Visual Representation Learning.} Action recognition requires accurately capturing temporal semantic variations. Numerous studies~\cite{zhu2023stmt, yang2022recurring, wang2021action, thatipelli2022strm, wang2021actionclip} have thus focused on visual representation learning. Early works studied the joint learning of spatial and temporal features of a video using various architectures~\cite{ni2022expanding, videoswin_22_cvpr, arnab2021vivit, ryoo2021tokenlearner}. Benefiting from large-scale visual and visual-text pre-training, recent methods leverage the strong spatial features learned in pre-training and focus on fine-tuning the trained model to capture the temporal semantics of a video.
However, these methods do not explicitly address the fundamental challenge of recognizing video actions from ambiguous and non-uniform videos. This paper proposes decomposing video actions into atomic actions to improve the recognition of challenging videos.

\vspace{+0.5mm}
\noindent{\bf Visual-text Representation Learning.} 
Several methods~\cite{chen2023FDT, wang2023seeing, wasim2023vita, wu2023revisiting, wu2023bidirectional} have been developed to overcome these limitations by identifying video frames that strongly align with the text prompts of video actions for video action recognition in vision-language models, \eg, CLIP. One remarkable work is BIKE~\cite{wu2023bidirectional}, which weights the video frames based on their alignment with the text prompts to compute a video embedding for action recognition. However, these methods are limited by the granularity difference between the video frames and the text prompts, where the text prompt is a global context for the video. Unlike previous methods, we generate sub-texts from the global text to describe atomic actions in video frames. The sub-texts are then used to construct coarse and fine-grained video embeddings, which improves the recognition of fine-grained actions.

\section{Methodology}
\label{Method}

\vspace{+0.5mm}
\noindent{\bf Preliminary.}
CLIP \cite{clip}, a visual-language pre-training method, consists of a visual encoder and a textual encoder. It learns a joint embedding space by maximizing the embedding similarities between aligned image-text pairs and minimizing the similarities for misaligned pairs. 
Given an image $\mathbf{I}$, and global text prompts $\{\mathbf{T}_{\text{c}}\}_{\text{c}=1}^{C}$ of $C$ classes formatted as \texttt{[a photo of a {class}]} that globally describe class semantics, where \texttt{{class}} represents the class name, CLIP performs zero-shot object recognition by extracting the visual class embedding $\mathbf{v}^\text{cls}$ from $\mathbf{I}$ and the text class embeddings $\{\mathbf{t}^\text{cls}_{\text{c}}\}_{\text{c}=1}^{C}$ from $\{\mathbf{T}_{\text{c}}\}_{\text{c}=1}^{C}$ using the visual encoder and the textual encoder, respectively. The image $\mathbf{I}$ is then classified into the class $\text{c}'$ with the maximum cosine similarity, \ie, $\text{c}' = \arg \max_{\text{c}} \text{sim}(\mathbf{v}^\text{cls}, \mathbf{t}^\text{cls}_{\text{c}})$.

\vspace{+0.5mm}
\noindent{\bf Overview.}
To extend CLIP for classifying the video $\{\mathbf{I}_{\text{l}}\}_{\text{l}=1}^{L}$ with $L$ frames, we adaptively compute a coarse video embedding and a fine-grained video embedding. The two multi-granularity embeddings can spot global semantic and atomic semantics of ambiguous and non-uniform actions in the video, and for computing the cosine similarity with the text class embedding $\text{c}$.
There are three key steps: 
i) Sub-text generation, decomposing each global text prompt $\mathbf{T}_{\text{c}}$ into a sequence of $N$ atomic action text descriptions $\{\mathbf{S}_{\text{c}, \text{n}}\}_{\text{n}=1}^{N}$ (\ie, sub-text prompts), by using a pre-trained large language model; 
ii) Coarse video embedding, embedding each video frame $\mathbf{I}_{\text{l}}$ into a visual class embedding $\mathbf{v}_{\text{l}}^\text{cls}$ that is a frame embedding. Then, by using the global text prompt augmented with sub-text prompts to identify salient video frames in CLIP embedding space, we aggregate the frame embedding $\{\mathbf{v}_{\text{l}}^\text{cls}\}_{\text{l}=1}^{L}$ into coarse video embeddings $\mathbf{o}^\text{coarse}$; 
iii) Fine-grained video embedding, finding video frames with atomic actions by using sub-text prompts and CLIP for computing a fine-grained video embedding $\mathbf{o}^\text{fine}$. We fuse the coarse and fine-grained video embedding into a video embedding $\mathbf{o}$ to compute cosine similarities with text class embeddings $\mathbf{t}^\text{cls}_{\text{c}}$ for classifying the video. The overview of our method is in Fig.~\ref{fig:framework}.

\subsection{Sub-text Generation}
\label{subsec:Sub_Generation}
A global text prompt $\mathbf{T}_{\text{c}}$ is a description of a video action. It overlooks that an action is usually formed by performing a sequence of atomic actions, like a video performing the action. Directly aligning video frame embedding to the global text prompt $\mathbf{T}_{\text{c}}$ introduces the granularity differences and decreases the video classification performance. To minimize human burden and bias, we propose a pipeline that automatically generates and selects sub-texts based on the global text (class name).

We leverage a pre-trained large language model $\text{LLM}(\cdot)$ to decompose global text prompt $\mathbf{T}_{\text{c}}$ into $N$ potentially atomic actions $\{\mathbf{S}_{\text{c}, \text{n}}\}_{\text{n}=1}^{N}$ by 
\begin{align}
    \{\mathbf{S}_{\text{c}, \text{n}}\}_{\text{n}=1}^{N} = \text{LLM}(\mathbf{P}_{\text{c}}) \ ,
\end{align}
where $\mathbf{P}_{\text{c}}$ is the prompt for describing the action class $\text{c}$ and directing the pre-trained large language model to find potential atomic action descriptions, sub-texts  $\{\mathbf{S}_{\text{c}, \text{n}}\}_{\text{n}=1}^{N}$. Refer to Fig.~\ref{fig:prompts_1} for an example. 

However, the existence of various pre-trained large language models and the flexibility in designing $\mathbf{P}_{\text{c}}$ make selecting a meaningful sub-text set challenging. We assume that an optimal sub-text set $\{\mathbf{S}_{\text{c}, \text{n}}\}_{\text{n}=1}^{N}$ for video classification should be sufficiently related with the global text prompt $\mathbf{T}_{\text{c}}$ while each sub-text should be diverse from the others. To measure the similarities and diversity, we propose a text prompt perplexity metric to select the sub-text set with the highest text prompt perplexity score. 

With the class embedding of the global text $\mathbf{T}_{\text{c}}$ and sub-texts $\{\mathbf{S}_{\text{c}, \text{n}}\}_{\text{n}=1}^{N}$ from the CLIP textual encoder, $\mathbf{t}_{\text{c}}^\text{cls}$ and $\mathbf{s}_{\text{c},\text{n}}^\text{cls}$, we define the text prompt perplexity score $\text{TPP}_\text{c}$ for  $\mathbf{t}_{\text{c}}^\text{cls}$ and $\mathbf{s}_{\text{c},\text{n}}^\text{cls}$ as below
\begin{align}
    &\text{TPP}_\text{c} = \exp \bigg( - \frac{1}{N} \sum_{\text{n}=1}^{N} \log \big( \alpha (\sigma_{\text{c},\text{n}} ) \beta(\delta_{\text{c},\text{n}}) \big) \bigg) \ ,
\end{align}
\begin{align}
    \sigma_{\text{c},\text{n}} = \frac{\text{sim}(\mathbf{t}_{\text{c}}^\text{cls}, \mathbf{s}_{\text{c},\text{n}}^\text{cls})+1}{2}  \ , 
\end{align}
\begin{align}
    \delta_{\text{c},\text{n}} =  1 - \frac{1}{N-1} \sum_{\text{n}'=1, \text{n}'\neq \text{n}}^{N} \frac{\text{sim}(\mathbf{s}_{\text{c},\text{n}}^\text{cls}, \mathbf{s}_{\text{c},\text{n}^{\prime}}^\text{cls})+1}{2}  \ ,
\end{align}
where {the $\mathrm{sim}(\cdot,\cdot)$ calculates the cosine similarities of inputs}, and $\sigma_{\text{c},\text{n}}$ and $\delta_{\text{c},\text{n}}$ denote the normalized similarity and divergence scores, respectively. Here, $\alpha(\cdot)$ and $\beta(\cdot)$ are scaling functions for the scores.  Please refer to Fig.~\ref{fig:ablation:TPP_performance} for the correlations between the video classification performance and the text prompt perplexity score $\text{TPP}_\text{c}$. 

\begin{table}[t]
    \centering
    \captionsetup{type=figure}
    \begin{subtable}[t]{\linewidth}
    \centering
    \begin{tikzpicture}
        \node (img) {\begin{tabular}{@{}c@{}c@{}c@{}c@{}c@{}c@{}}
            \hspace{-2.0cm}
            \raisebox{-1cm}{
            \rotatebox{90}{\parbox[c][0.5cm][c]{1.5cm}{\centering\small\scalebox{0.8}{Springboard diving}}}}&
            \hspace{-0.25cm}
            \multirow{2}{*}{
            \begin{minipage}[c][2.6cm]{1cm}
                \hspace{-0.0cm}\raisebox{2.9cm}{\includegraphics[width=1.1cm, height=1.1cm]{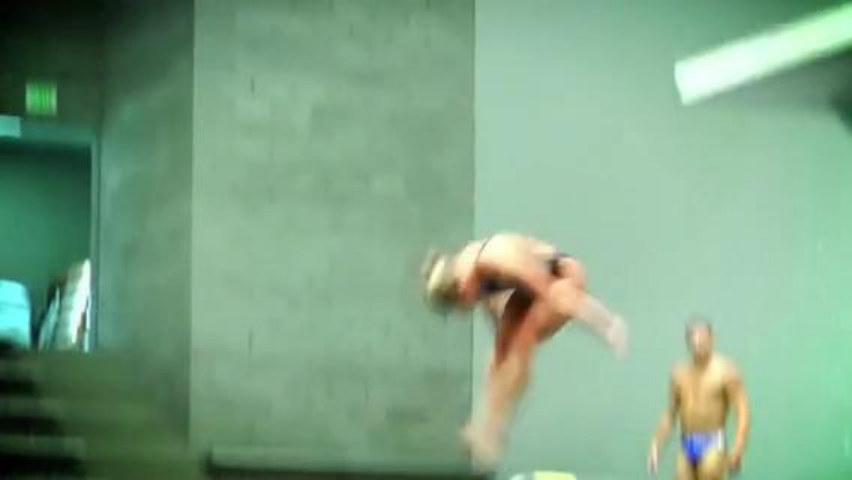}}\\
                [-14.8em]
                \raisebox{2.1cm}{\includegraphics[width=1.1cm, height=1.1cm]{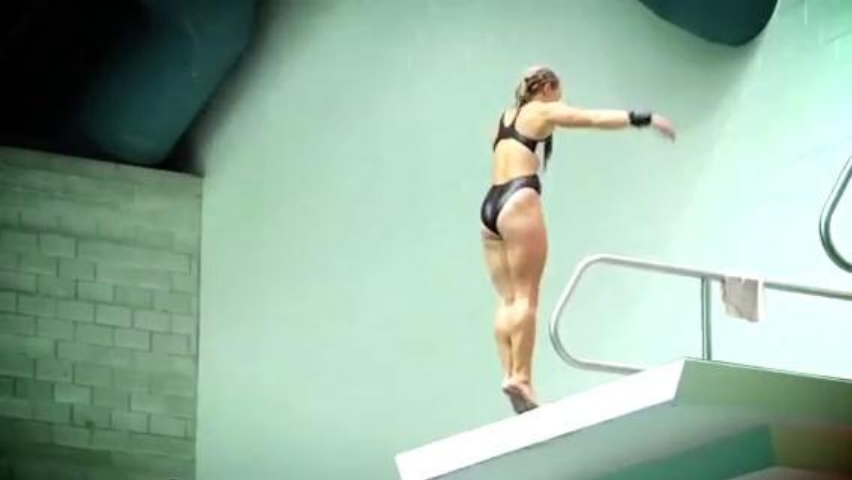}}
            \end{minipage}
            }&
            \hspace{-0.1cm}
            \multirow{2}{*}{
            \begin{minipage}[c][2.6cm]{1cm}
                \hspace{-0.0cm}\raisebox{2.9cm}{\includegraphics[width=1.1cm, height=1.1cm]{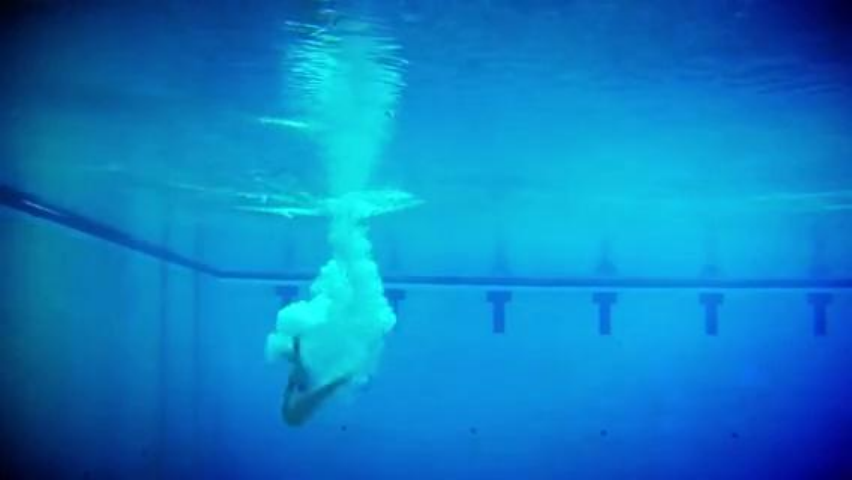}}\\
                [-14.8em]
                \raisebox{2.1cm}{\includegraphics[width=1.1cm, height=1.1cm]{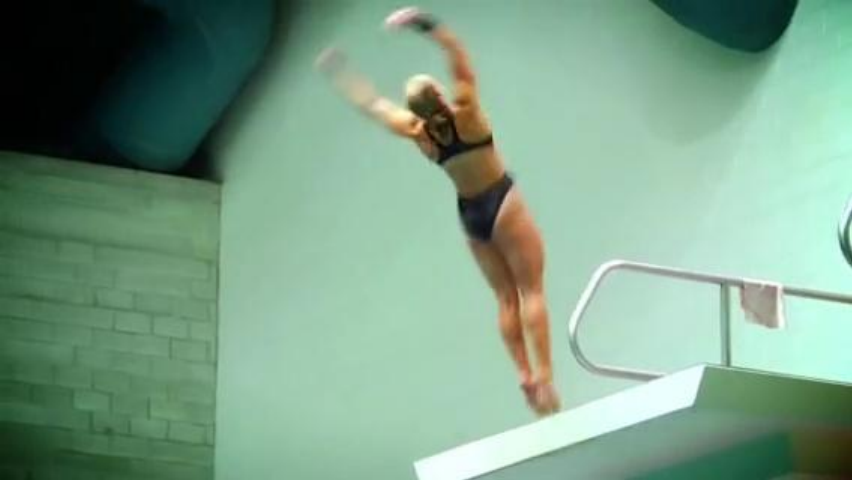}}
            \end{minipage}
            }&
            \hspace{-0.4cm}
            \raisebox{-0.08cm}{
              \parbox[c]{1cm}{
                \tiny
                \centering
                \tikz[baseline]{\node[fill=yellow!50,minimum width=0.5cm,minimum height=1.1cm, inner sep=0pt] {\rotatebox{90}{\scalebox{0.9}{\small\text{GPT-3.5}}}};}
                \\[0.2em]
                \centering
                \tikz[baseline]{\node[fill=yellow!50,minimum width=0.5cm,minimum height=1.1cm, inner sep=0pt] {\rotatebox{90}{\scalebox{0.9}{\small\text{LLaVa1.6}}}};}
                \\[-0.0em]
              }
            }& 
            \hspace{-0.4cm}
            \raisebox{-0.05cm}{
              \parbox[l]{1cm}{
                \raggedright
                \begin{tikzpicture}[baseline]
                  \node[minimum width=2.1cm, minimum height=0.27cm, inner sep=0pt, text width=2.1cm, align=left] {
                    \scalebox{0.9}{\text{1. Preparing}}
                  };
                \end{tikzpicture}\\[0.2em]
                \begin{tikzpicture}[baseline]
                  \node[minimum width=2.1cm, minimum height=0.27cm, inner sep=0pt, text width=2.1cm, align=left] {
                    \scalebox{0.9}{\text{3. Rotation}}
                  };
                \end{tikzpicture}\\[0.15em]
                \rule{2.85cm}{0.2pt}\\[-0.1em]
                \begin{tikzpicture}[baseline]
                  \node[minimum width=2.1cm, minimum height=0.27cm, inner sep=0pt, text width=2.1cm, align=left] {
                    \scalebox{0.9}{\text{1. Prepare diving}}
                  };
                \end{tikzpicture}\\[0.2em]
                \begin{tikzpicture}[baseline]
                  \node[minimum width=2.1cm, minimum height=0.27cm, inner sep=0pt, text width=2.1cm, align=left] {
                    \scalebox{0.9}{\text{3. In mid-air}}
                  };
                \end{tikzpicture}\\[-0.15em]
              }
            }&
            \hspace{1.17cm}
            \raisebox{-0.1cm}{
              \parbox[l]{1cm}{
                \raggedright
                \begin{tikzpicture}[baseline]
                  \node[minimum width=2.1cm, minimum height=0.27cm, inner sep=0pt, text width=2.1cm, align=left] {
                    \scalebox{0.9}{\text{2. Takeoff}}
                  };
                \end{tikzpicture}\\[0.2em]
                \begin{tikzpicture}[baseline]
                  \node[minimum width=2.1cm, minimum height=0.27cm, inner sep=0pt, text width=2.1cm, align=left] {
                    \scalebox{0.9}{\text{4. Entry the water}}
                  };
                \end{tikzpicture}\\[0.183em]
                \rule{2.9cm}{0.2pt}\\[-0.1em]
                \begin{tikzpicture}[baseline]
                  \node[minimum width=2.1cm, minimum height=0.27cm, inner sep=0pt, text width=2.1cm, align=left] {
                    \scalebox{0.9}{\text{2. Jump off the board}}
                  };
                \end{tikzpicture}\\[0.2em]
                \begin{tikzpicture}[baseline]
                  \node[minimum width=2.1cm, minimum height=0.27cm, inner sep=0pt, text width=2.1cm, align=left] {
                    \scalebox{0.9}{\text{4. Dive into the water}}
                  };
                \end{tikzpicture}\\[0.15em]
              }
            }
            \end{tabular}};
        \end{tikzpicture}
        \vspace{-0.25cm}
    \end{subtable}
    \vspace{-0.7cm}
    \caption{\small \it Example sub-texts generated from GPT-3.5 and LLaVa1.6.
    }
    \label{fig:prompts_1}
\end{table}
\begin{table}[t]
    \centering
    \captionsetup{type=figure}
    \begin{subtable}[t]{\linewidth} 
        \centering
        \begin{tikzpicture}
            \node (img) {\begin{tabular}{@{}c@{}c@{}c@{}c@{}c@{}}
                \hspace{-0.35cm}
                    \makebox[0.5cm][c]{
            \raisebox{-0.3cm}{
                \rotatebox{90}{
                    \parbox[c][1.0cm][c]{0.5cm}{
                        \centering
                        \small\scalebox{0.8}{(a) Eating cake}
                    }
                }
            }
        }&  
                \hspace{-0.10cm}\includegraphics[width=2.01cm, height=1.2cm]{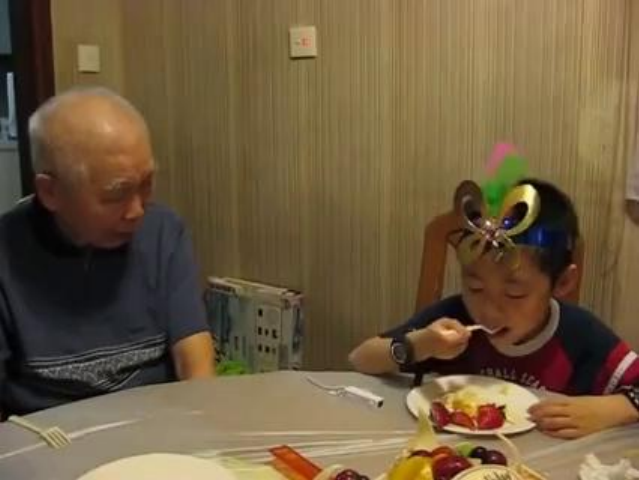} 
                & 
                \hspace{0.05cm}\includegraphics[width=2.01cm, height=1.2cm]{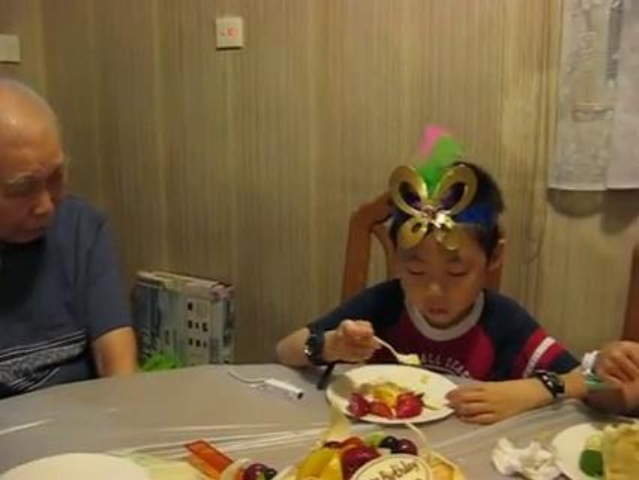}
                &
                \hspace{0.05cm}\includegraphics[width=2.01cm, height=1.2cm]{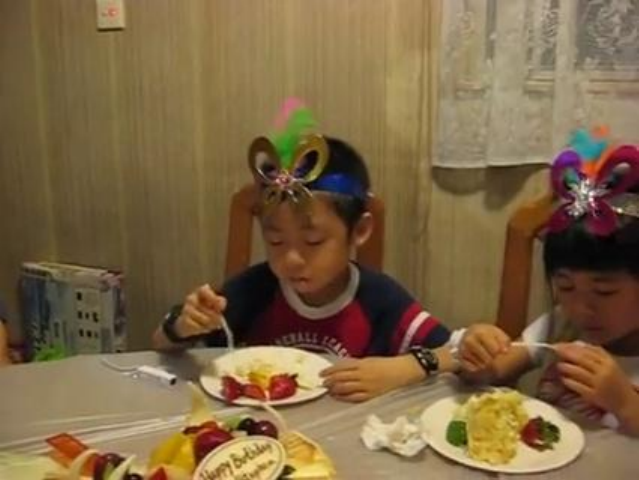}
                &
                \hspace{0.05cm}\includegraphics[width=2.01cm, height=1.2cm]{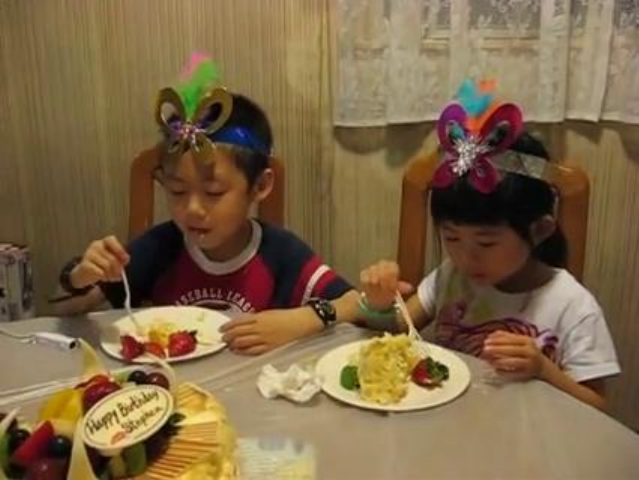}
            \end{tabular}};
        \end{tikzpicture}
        \vspace{-1.25cm}
        \begin{flushleft}
            \raisebox{0cm}{
                \begin{tabular}{cccc}
                    \multicolumn{1}{c}{\hspace{-0.5cm}\parbox[c][0.5cm][c]{0.33\linewidth}{\centering\scalebox{0.8}{\small\text{(b) Tasting food}}}}&
                    \multicolumn{1}{c}{\hspace{-1.1cm}\parbox[c][0.5cm][c]{0.33\linewidth}{\centering\scalebox{0.8}{\small\text{(c) Dining}}}}&
                    \multicolumn{1}{c}{\hspace{-1.15cm}\parbox[c][0.5cm][c]{0.33\linewidth}{\centering\scalebox{0.8}{\small\text{(d) Eating spaghetti}}}}&
                    \multicolumn{1}{c}{\hspace{-1.11cm}\parbox[c][0.5cm][c]{0.33\linewidth}{\centering\scalebox{0.8}{\small\text{(e) Eating burger}}}}
                \end{tabular}}
        \end{flushleft}
        \vspace{-0.25cm}
    \end{subtable}
    \begin{subtable}[t]{\linewidth} 
        \centering
        \begin{tikzpicture}
            \node (img) {\begin{tabular}{@{}c@{}c@{}}
                \vspace{-0.15cm}
                \hspace{-0.7cm}
                \begin{tikzpicture}[baseline,scale=0.7]
                    \begin{axis}[
                        title={Single},
                        ybar,
                        grid,
                        symbolic x coords={(a), (b), (c), (d), (e)},
                        xtick=data,
                        ylabel=\empty,
                        xlabel={Categories},
                        xlabel=\empty,
                        y tick label style={xshift=0cm},
                        x tick label style={yshift=0cm},
                        ymin=0,
                        ymax=30,
                        bar width=15pt,
                        nodes near coords,
                        width=0.8\textwidth,
                        height=0.5\textwidth,
                        enlarge x limits=0.15,
                        enlarge y limits=0.1,
                        ylabel style={yshift=-0.95em},
                        x label style={at={(axis description cs:0.5,0.1)}, anchor=north},
                    ]
                        \addplot [forestgreen, fill=forestgreen!30!white]coordinates {((a),23) ((b),28) ((c),25) ((d),20) ((e),4)};
                    \end{axis}
                \end{tikzpicture}
                &
                \hspace{0.10cm}
                \begin{tikzpicture}[baseline,scale=0.7]
                    \begin{axis}[
                        title={Coarse video embedding},
                        ybar,
                        grid,
                        symbolic x coords={(a), (b), (c), (d), (e)},
                        xtick=data,
                        ylabel = \empty,
                        xlabel={Categories},
                        xlabel=\empty,
                        y tick label style={xshift=0cm},
                        x tick label style={yshift=0cm},
                        ymin=0,
                        ymax=30,
                        bar width=15pt,
                        nodes near coords,
                        width=0.8\textwidth,
                        height=0.5\textwidth,
                        enlarge x limits=0.15,
                        enlarge y limits=0.1,
                        ylabel style={yshift=-0.95em},
                        x label style={at={(axis description cs:0.5,0.1)}, anchor=north},
                    ]
                        \addplot [blue, fill=blue!30!white]coordinates {((a),28) ((b),26) ((c),26) ((d),17) ((e),3)};
                    \end{axis}
                \end{tikzpicture}
                \end{tabular}};
        \end{tikzpicture} 
    \end{subtable}
    \vspace{-0.7cm}
    \caption{\it \small 
    An example illustrating the benefits of using our coarse video embedding for ambiguous actions. We show a video in the first row, with the ground truth action labeled as (a) on the left. Its ambiguous actions are annotated with (b), (c), (d) and (e) at the bottom of the video. The action probability from the single granularity-based method \cite{wu2023bidirectional} and our coarse video embedding are in the next row. Compared with the single granularity-based method that identifies the video as (b) `Tasting food', we obtain a higher probability of the correct class (a) `Eating cake'. 
    }
    \label{fig:weight_1}
\end{table}

\begin{table*}[t]
    \centering
    \caption{\it \small Comparisons results on the Kinetics-400 dataset. We report the FLOPs in inference phase.``Views'' indicates \# temporal clip $\times$ \# spatial crop. The magnitude is Million ($10^{6}$) for parameters (Param). We achieve the highest top-1 and top-5 accuracy by employing only an 8-frame input evaluated with a single view. We highlight the best numbers in \textbf{\textit{bold}}
.}
    \hspace{-0.2cm} 
    \scalebox{1}{
    \resizebox{\linewidth}{!}{%
  \begin{tabular}{lcccccccc}
\toprule
  \textbf{Method} & \textbf{Venue} & \textbf{Input}   & \textbf{Pre-training} & \textbf{Top-1(\%)} & \textbf{Top-5(\%)} &  \textbf{Views} & \textbf{GFLOPs} & \textbf{Param}  \\
  \hline
  \rowcolor{gray!20}
 \multicolumn{9}{l}{\emph{Methods with large-scale visual pre-training}} \\
  MVFNet$_{En}$~\cite{wu2021mvfnet} & AAAI'21 & 24\x224$^2$  & ImageNet-1K & 79.1 & 93.8 & 10\x3 & 188\x30 & - \\
  ActionCLIP ViT-B/16~\cite{wang2021actionclip} & arXiv'21  & 16\x224$^2$  & WIT-400M & 82.6 & 96.2 & 10\x3 & 282\x30 & 141.7 \\ 
  ViViT-B/16\x2~\cite{arnab2021vivit} & ICCV'21 & 32\x224$^2$ & JFT-300M & 80.0 & - & 4\x1 & 455.2 & 151.9\\
  ST-Adapter ViT-B/16~\cite{pan2022st} & NeurIPS'22 & 8\x224$^2$  & WIT-400M & 82.0 & 95.7 & 3\x1 & 455 & 128.8 \\
  EVL ViT-B/16~\cite{lin2022frozen} & ECCV'22 & 8\x224$^2$  & WIT-400M & 82.9 & - & 3\x1 & 444 & 177.7 \\
  MTV-B~\cite{yan2022multiview} & CVPR'22 & 32\x224$^2$   & JFT-300M & 81.8 & 95.0 & 4\x3 & 384\x12 & 310 \\
  VideoSwin-B~\cite{videoswin_22_cvpr} & CVPR'22 & 32\x224$^2$  & ImageNet-21K & 82.7 & 95.5 & 4\x3 & 282\x12 & 88.1 \\ 
  ATM ViT-B/16~\cite{atm} & ICCV'23 & 8\x224$^2$ & WIT-400M & 82.8 & 95.6 & 1\x1 & 378\x1 & -\\
  DIST ViT-B/16~\cite{qing2023disentangling}& ICCV'23 & 8\x224$^2$  & WIT-400M & 83.6 & - & 3\x1 & 163\x3 & 105 \\
  AIM ViT-B/16~\cite{yang2023aim} & ICLR'23 & 8\x224$^2$ & WIT-400M & 83.9 & 96.3 & 3\x1 & 202\x3 & 97 \\  
  ILA-ViT-B/16~\cite{tu2023implicit} &ICCV'23 & 8\x224$^2$ & WIT-400M &
  84.0 & 96.6 & 4\x3 & 149\x12 & - \\ 
  \hline
 \rowcolor{gray!20}
 \multicolumn{9}{l}{\emph{Methods with large-scale visual-text pre-training}} \\
  X-CLIP ViT-B/16~\cite{ni2022expanding} & ECCV'22 & 8\x224$^2$  & WIT-400M & 82.3 & - & 1\x1 & 145\x1 & - \\
  VideoPrompt ViT-B/16~\cite{ju2022prompting} & ECCV'22 & 16\x224$^2$  & WIT-400M & 76.9 & 93.5 & 1\x5 & - & 154 \\ 
  SIF ViT-B/16~\cite{wang2023seeing} & ACMMM'23 & 8\x224$^2$  & WIT-400M & 77.4 & 93.6 & 4\x3 & 1136\x12 & 143.9 \\
  Vita-CLIP ViT-B/16~~\cite{wasim2023vita} & CVPR'23 & 8\x224$^2$  & WIT-400M & 80.5 & 96.0 & 1\x1 & 97\x1 & 187.9 \\
  Vita-CLIP ViT-B/16~\cite{wasim2023vita} & CVPR'23 & 8\x224$^2$  & WIT-400M & 81.8 & 96.0 & 4\x3 & 97\x12 & 187.9 \\
  BIKE ViT-B/16~\cite{wu2023bidirectional} & CVPR'23 & 8\x224$^2$  & WIT-400M & 83.2 & - & 1\x1 & - & 124.1 \\
  BIKE ViT-B/16~\cite{wu2023bidirectional} & CVPR'23 & 8\x224$^2$  & WIT-400M & 83.9 & - & 4\x3 & - & 124.1 \\
  M$^2$-CLIP ViT-B/16~\cite{wang2024multimodal} & AAAI'24 & 8\x224$^2$  & WIT-400M & 83.4 & 96.3 & 4\x3 & 214\x3 & 165 \\
  \midrule
  \multirow{1}{*}{\textbf{Ours~ViT-B/16}} & \multirow{1}{*}{-} & 8\x224$^2$  & \multirow{1}{*}{WIT-400M} & \textbf{84.5} & \textbf{96.7} & \textbf{1\x1} & 90.2\x1 & 126.1 \\ 
\bottomrule
  \end{tabular}}} 
    \label{tab:k400_sota}
\end{table*}

\subsection{Coarse Video Embedding}
For identifying different atomic actions in ambiguous videos, we augment the global text $\mathbf{T}_{\text{c}}$ with the sub-texts $\{\mathbf{S}_{\text{c}, \text{n}}\}_{\text{n}=1}^{N}$ to adaptively aggregate the video frames $\{\mathbf{I}_{\text{l}}\}_{\text{l}=1}^{L}$ in CLIP embedding space. As the class name in global text is usually a phrase, and sub-text $\mathbf{S}_{\text{c}, \text{n}}$ describes the atomic action with multiple words, we keep the word embeddings to benefit from the rich semantics of each word embedding for video frame aggregations. We denote the matrix form of the word embeddings for global text and sub-text as $\mathcal{T}_{\text{c}}$ and $\mathcal{S}_{\text{c}, \text{n}}$, and augment $\mathcal{T}_{\text{c}}$ by the cross-attention mechanism, 
\begin{align}
    &\mathcal{Q}_{\text{c}} = \mathcal{T}_{\text{c}} \mathcal{W}^{\text{q}} \ , \quad  \mathcal{K}_{\text{c}, \text{n}} = \mathcal{S}_{\text{c}, \text{n}} \mathcal{W}^{\text{k}} \ , \quad \mathcal{V}_{\text{c}, \text{n}} = \mathcal{S}_{\text{c}, \text{n}} \mathcal{W}^{\text{v}} \ ,  \\
    &\mathcal{K}_{\text{c}} = [\{\mathcal{K}_{\text{c}, \text{n}}\}_{\text{n}=1}^{N}] \ , \quad \mathcal{V}_{\text{c}} = [\{\mathcal{V}_{\text{c}, \text{n}}\}_{\text{n}=1}^{N}] \ , \\
    &\hat{\mathcal{T}}_{\text{c}} = \text{Attention}( \mathcal{Q}_{\text{c}}, \mathcal{K}_{\text{c}}, \mathcal{V}_{\text{c}}) + \mathcal{T}_{\text{c}} \ ,
\end{align}
where $\mathcal{W}^{\text{q}}$, $\mathcal{W}^{\text{k}}$, and $\mathcal{W}^{\text{v}}$ are matrices for projecting the query $\mathcal{Q}_{\text{c}}$,  key $\mathcal{K}_{\text{c}, \text{n}}$, and value $\mathcal{V}_{\text{c}, \text{n}}$. Here, $[\{\mathcal{K}_{\text{c}, \text{n}}\}_{\text{n}=1}^{N}]$ and $[\{\mathcal{V}_{\text{c}, \text{n}}\}_{\text{n}=1}^{N}]$ concatenate all keys and values for the query $\mathcal{Q}_{\text{c}}$.


Using the augmented global text embeddings $\hat{\mathcal{T}}_{\text{c}}$, we find the salient video frame embedding $\mathbf{v}_{\text{l}}^\text{cls}$ for aggregating a coarse video embedding. To find the coarse-grained importance score $a_{\text{l}}^\text{coarse}$ of each frame embedding, we calculate the overall normalized similarity between each word embedding and the video frame embedding by
\begin{align}
    a_{\text{l}}^\text{coarse} = \sum_{\hat{\mathbf{t}}_{\text{c}} \in \hat{\mathcal{T}}_{\text{c}}} \frac{\exp\big(\text{sim}(\hat{\mathbf{t}}_{\text{c}},\mathbf{v}_{\text{l}}^\text{cls}) \big)}{\sum_{\text{l}'=1}^{L} \text{sim}(\hat{\mathbf{t}}_{\text{c}},\mathbf{v}_{\text{l}'}^\text{cls})} \ .
\end{align}
We then compute the coarse video embedding $\mathbf{o}^\text{coarse}$ with summing over the video frame embedding $\mathbf{v}_{\text{l}}^\text{cls}$ weighted by $a_{\text{l}}^\text{coarse}$, 
\begin{align}
    \mathbf{o}^\text{coarse} = \sum_{\text{l}=1}^{L} \mathbf{v}_{\text{l}}^\text{cls} a_{\text{l}}^\text{coarse} \ . \label{eq:coarse}
\end{align}
In Fig.~\ref{fig:weight_1}, we show amibugous action probability computed from a single granularity-based method \cite{wu2023bidirectional} and our coarse video embedding. Our method uses sub-text to augment the global text help the model to identify different atomic actions in ambiguous actions, and improve the accuracy of recognizing ambiguous actions.

\begin{table}[t]
    \centering
    \captionsetup{type=figure}
    \begin{subtable}[t]{\linewidth} 
        \centering
        \begin{tikzpicture}
            \node (img) {\begin{tabular}{ccccc}
                \hspace{-0.3cm}
                \makebox[0.5cm][c]{
                \raisebox{-0cm}{
                    \rotatebox{90}{
                        \parbox[c][2.0cm][c]{0.5cm}{
                            \centering
                            \small\scalebox{0.8}{Drinking}
                        }
                    }
                }
            }
                 & 
                 \hspace{-0.25cm}\includegraphics[width=1.84cm, height=1.05cm]{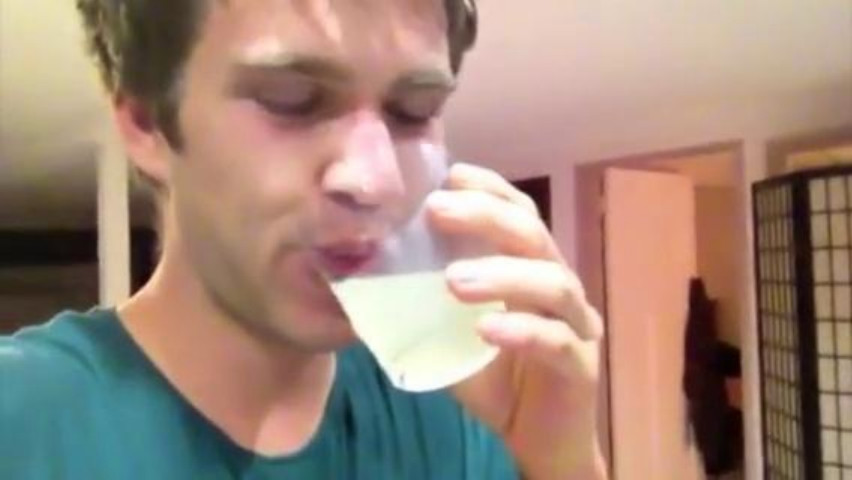} & \hspace{-0.4cm}\includegraphics[width=1.84cm, height=1.05cm]{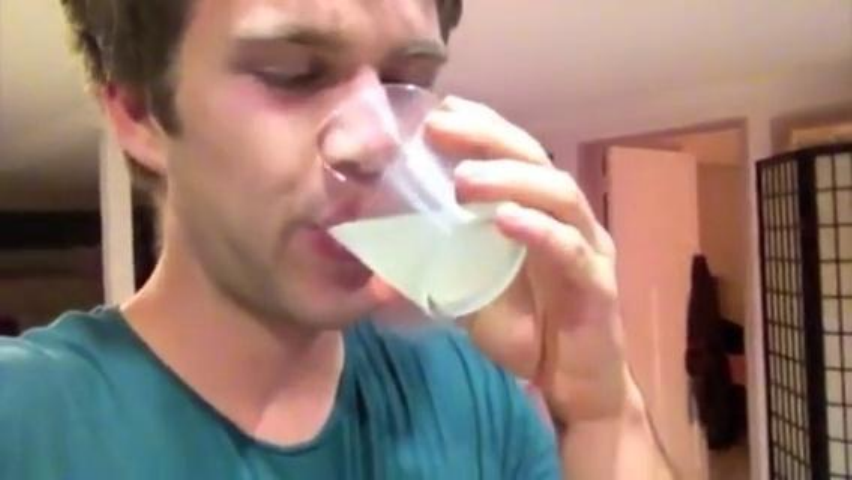} & \hspace{-0.4cm}\includegraphics[width=1.84cm, height=1.05cm]{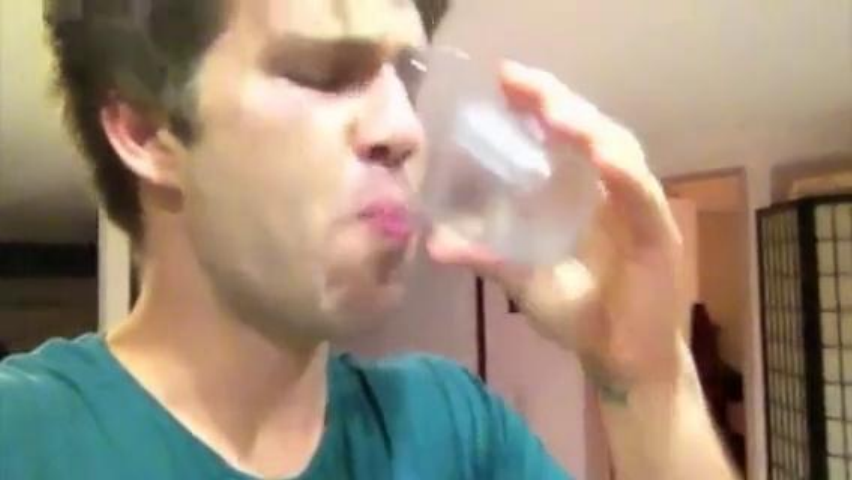} & \hspace{-0.4cm}\includegraphics[width=1.84cm, height=1.05cm]{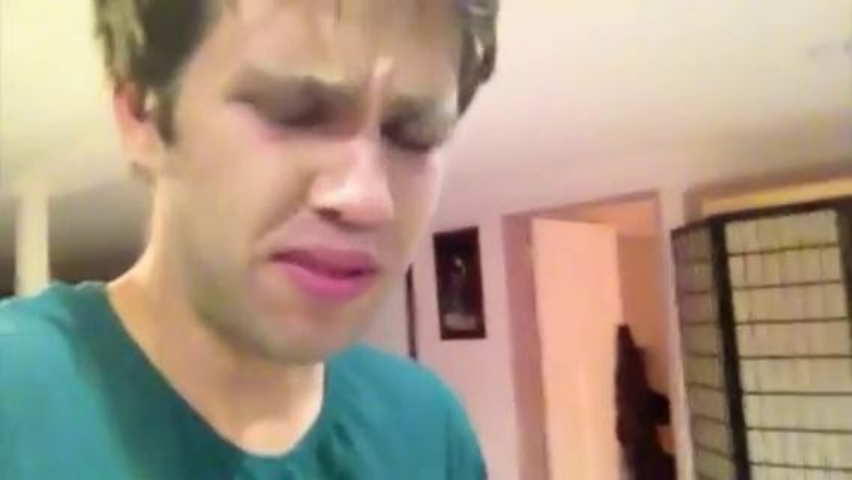}
            \end{tabular}};
        \end{tikzpicture}
        \vspace{-1.25cm}
        \begin{flushleft}
            \raisebox{0cm}{
                \begin{tabular}{ccccc}
                    \multicolumn{1}{c}{\hspace{-0.50cm}\parbox[c][0.5cm][c]{1.2cm}{\centering\scalebox{0.8}{\small $a_{\text{l}}^\text{coarse}$}}}&
                    \multicolumn{1}{c}{\hspace{-0.81cm}\parbox[c][0.5cm][c]{2.2cm}{\centering\tikz[baseline]{\node[fill=blue!65,minimum width=1.84cm,minimum height=0.20cm, inner sep=0pt] {\scalebox{0.8}{\small\text{23}}};}}}&
                    \multicolumn{1}{c}{\hspace{-0.71cm}\parbox[c][0.5cm][c]{2.1cm}{\centering\tikz[baseline]{\node[fill=blue!75,minimum width=1.84cm,minimum height=0.20cm, inner sep=0pt] {\scalebox{0.8}{\small\text{30}}};}}}&
                    \multicolumn{1}{c}{\hspace{-0.71cm}\parbox[c][0.5cm][c]{2.2cm}{\centering\tikz[baseline]{\node[fill=blue!67,minimum width=1.84cm,minimum height=0.20cm, inner sep=0pt] {\scalebox{0.8}{\small\text{24}}};}}}&
                    \multicolumn{1}{c}{\hspace{-0.75cm}\parbox[c][0.5cm][c]{2.2cm}{\centering\tikz[baseline]{\node[fill=blue!65,minimum width=1.84cm,minimum height=0.20cm, inner sep=0pt] {\scalebox{0.8}{\small\text{23}}};}}}
                \end{tabular}}
        \end{flushleft}
        \vspace{-0.85cm}
        \begin{flushleft}
            \raisebox{0cm}{
                \begin{tabular}{ccccc}
                    \multicolumn{1}{c}{\hspace{-0.50cm}\parbox[c][0.5cm][c]{1.2cm}{\centering\scalebox{0.8}{\small$a_{\text{l}}^\text{fine}$}}}&
                    \multicolumn{1}{c}{\hspace{-0.81cm}\parbox[c][0.5cm][c]{2.2cm}{\centering\tikz[baseline]{\node[fill=red!65,minimum width=1.84cm,minimum height=0.20cm, inner sep=0pt] {\scalebox{0.8}{\small\text{30}}};}}}&
                    \multicolumn{1}{c}{\hspace{-0.71cm}\parbox[c][0.5cm][c]{2.1cm}{\centering\tikz[baseline]{\node[fill=red!64,minimum width=1.84cm,minimum height=0.20cm, inner sep=0pt] {\scalebox{0.8}{\small\text{26}}};}}}&
                    \multicolumn{1}{c}{\hspace{-0.71cm}\parbox[c][0.5cm][c]{2.2cm}{\centering\tikz[baseline]{\node[fill=red!75,minimum width=1.84cm,minimum height=0.20cm, inner sep=0pt] {\scalebox{0.8}{\small\text{33}}};}}}
                    &
                    \multicolumn{1}{c}{\hspace{-0.75cm}\parbox[c][0.5cm][c]{2.2cm}{\centering\tikz[baseline]{\node[fill=red!35,minimum width=1.84cm,minimum height=0.20cm, inner sep=0pt] {\scalebox{0.8}{\small\text{11}}};}}}
                \end{tabular}}
        \end{flushleft}
        \vspace{-0.4cm}
    \end{subtable}
    \begin{subtable}[t]{\linewidth} 
        \centering
        \begin{tikzpicture}
            \node (img) {\begin{tabular}{ccccc}
                \hspace{-0.3cm}
                \makebox[0.5cm][c]{
                \raisebox{-0cm}{
                    \rotatebox{90}{
                        \parbox[c][2.0cm][c]{0.5cm}{
                            \centering
                            \small\scalebox{0.8}{ Clean} \\
                            \small\scalebox{0.8}{and jerk}
                        }
                    }
                }
            }
                 & 
                 \hspace{-0.25cm}\includegraphics[width=1.84cm, height=1.05cm]{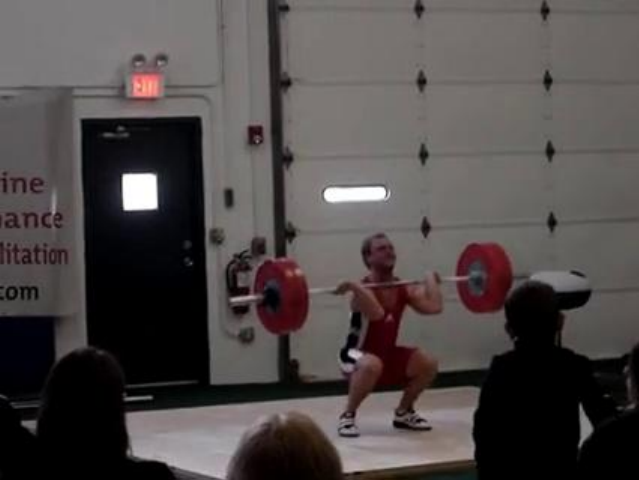} & \hspace{-0.4cm}\includegraphics[width=1.84cm, height=1.05cm]{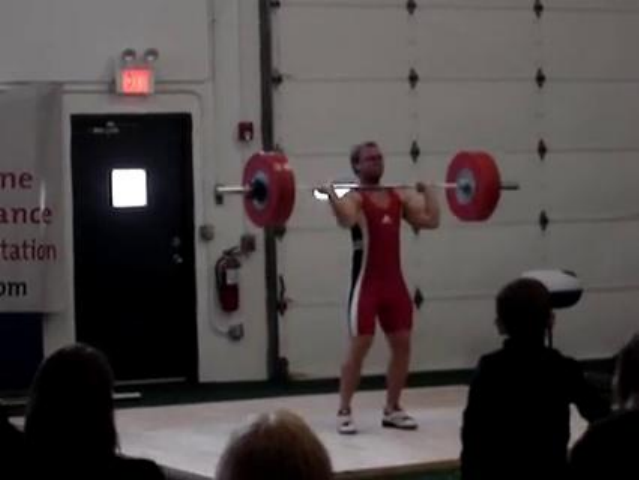} & \hspace{-0.4cm}\includegraphics[width=1.84cm, height=1.05cm]{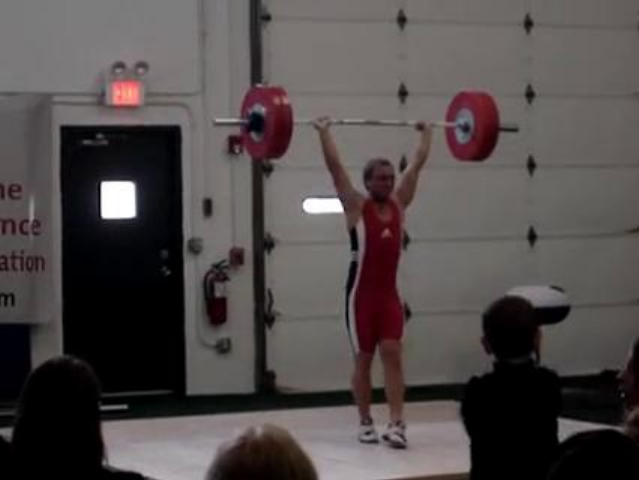} & \hspace{-0.4cm}\includegraphics[width=1.84cm, height=1.05cm]{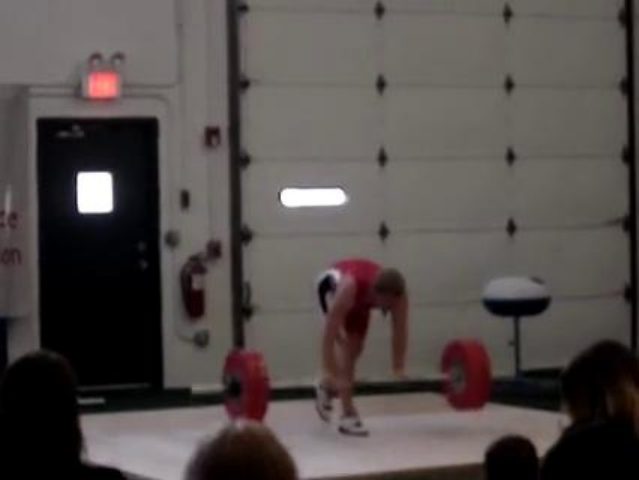}
            \end{tabular}};
        \end{tikzpicture}
        \vspace{-1.25cm}
        \begin{flushleft}
            \raisebox{0cm}{
                \begin{tabular}{ccccc}
                    \multicolumn{1}{c}{\hspace{-0.50cm}\parbox[c][0.5cm][c]{1.2cm}{\centering\scalebox{0.8}{\small $a_{\text{l}}^\text{coarse}$}}}&
                    \multicolumn{1}{c}{\hspace{-0.81cm}\parbox[c][0.5cm][c]{2.2cm}{\centering\tikz[baseline]{\node[fill=blue!75,minimum width=1.84cm,minimum height=0.20cm, inner sep=0pt] {\scalebox{0.8}{\small\text{56}}};}}}&
                    \multicolumn{1}{c}{\hspace{-0.71cm}\parbox[c][0.5cm][c]{2.1cm}{\centering\tikz[baseline]{\node[fill=blue!60,minimum width=1.84cm,minimum height=0.20cm, inner sep=0pt] {\scalebox{0.8}{\small\text{22}}};}}}&
                    \multicolumn{1}{c}{\hspace{-0.71cm}\parbox[c][0.5cm][c]{2.2cm}{\centering\tikz[baseline]{\node[fill=blue!45,minimum width=1.84cm,minimum height=0.20cm, inner sep=0pt] {\scalebox{0.8}{\small\text{13}}};}}}&
                    \multicolumn{1}{c}{\hspace{-0.75cm}\parbox[c][0.5cm][c]{2.2cm}{\centering\tikz[baseline]{\node[fill=blue!30,minimum width=1.84cm,minimum height=0.20cm, inner sep=0pt] {\scalebox{0.8}{\small\text{9}}};}}}
                \end{tabular}}
        \end{flushleft}
        \vspace{-0.85cm}
        \begin{flushleft}
            \raisebox{0cm}{
                \begin{tabular}{ccccc}
                    \multicolumn{1}{c}{\hspace{-0.50cm}\parbox[c][0.5cm][c]{1.2cm}{\centering\scalebox{0.8}{\small$a_{\text{l}}^\text{fine}$}}}&
                    \multicolumn{1}{c}{\hspace{-0.81cm}\parbox[c][0.5cm][c]{2.2cm}{\centering\tikz[baseline]{\node[fill=red!75,minimum width=1.84cm,minimum height=0.20cm, inner sep=0pt] {\scalebox{0.8}{\small\text{27}}};}}}&
                    \multicolumn{1}{c}{\hspace{-0.71cm}\parbox[c][0.5cm][c]{2.1cm}{\centering\tikz[baseline]{\node[fill=red!72,minimum width=1.84cm,minimum height=0.20cm, inner sep=0pt] {\scalebox{0.8}{\small\text{25}}};}}}&
                    \multicolumn{1}{c}{\hspace{-0.71cm}\parbox[c][0.5cm][c]{2.2cm}{\centering\tikz[baseline]{\node[fill=red!72,minimum width=1.84cm,minimum height=0.20cm, inner sep=0pt] {\scalebox{0.8}{\small\text{25}}};}}}
                    &
                    \multicolumn{1}{c}{\hspace{-0.75cm}\parbox[c][0.5cm][c]{2.2cm}{\centering\tikz[baseline]{\node[fill=red!70,minimum width=1.84cm,minimum height=0.20cm, inner sep=0pt] {\scalebox{0.8}{\small\text{23}}};}}}
                \end{tabular}}
        \end{flushleft}
    \end{subtable}
    \caption{\it \small An illustration of coarse and fine-grained importance score. The fine-grained importance scores \textnormal{$a_{\text{l}}^\text{fine}$} allocate larger scores than the coarse importance score \textnormal{$a_{\text{l}}^\text{coarse}$} to key frames. For example, the fourth frame in the second row `lowering barbell back down to the ground' represents an atomic action of `Clean and jerk', and \textnormal{$a_{\text{l}}^\text{fine}$} has a larger score than \textnormal{$a_{\text{l}}^\text{coarse}$}.
    }
    \label{fig:weight}
\end{table}

\subsection{Fine-grained Video Embedding}
While the video coarse embedding $\mathbf{o}^\text{coarse}$ captures salient semantics for ambiguous actions, it may overlook atomic actions in the video that align with the expected action class $\text{c}$ due to the non-uniform property of an actions. To address the limitation, we use the word embedding of the sub-texts to find atomic actions among the video frame embeddings $\{\mathbf{v}_{\text{l}}^\text{cls}\}_{l=1}^{L}$ for computing the fine-grained video embedding.

 We compute a fine-grained importance score $a_{\text{l}}^\text{fine}$ for each video frame embedding $\mathbf{v}_{\text{l}}^\text{cls}$ as the maximum average similarity between $\mathbf{v}_{\text{l}}^\text{cls}$ and $\mathcal{S}_{\text{c}, \text{n}}$, and normalize it by a SoftMax function,
\begin{align}
    a_{\text{l}}^\text{fine} =  \frac{\exp \big( \max_{\text{n} }\frac{1}{|\mathcal{S}_{\text{c}, \text{n}}|} \sum{\mathbf{s}_{\text{c}, \text{n}} \in \mathcal{S}_{\text{c}, \text{n}}} \text{sim}(\mathbf{s}_{\text{c}, \text{n}},\mathbf{v}_{\text{l}}^\text{cls}) \big)}{\sum_{\text{l}'=1}^{L} \exp \big( \max_{\text{n} }\frac{1}{|\mathcal{S}_{\text{c}, \text{n}}|} \sum{\mathbf{s}_{\text{c}, \text{n}} \in \mathcal{S}_{\text{c}, \text{n}}} \text{sim}(\mathbf{s}_{\text{c}, \text{n}},\mathbf{v}_{\text{l}'}^\text{cls}) \big)} \ ,
\end{align}
where $|\mathcal{S}_{\text{c}, \text{n}}|$ is the number of words in the sub-text, and $\mathbf{s}_{\text{c}, \text{n}} \in \mathcal{S}_{\text{c}, \text{n}}$ is a word embedding  for the sub-text. Similar to Eq.~\eqref{eq:coarse}, we weight the video frame embedding $\mathbf{v}_{\text{l}}^\text{cls}$ with $a_{\text{l}}^\text{fine}$, and get the fine-grained video embedding $\mathbf{o}^\text{fine}$ by 
\begin{align}
    \mathbf{o}^\text{fine} = \sum_{\text{l}=1}^{L} \mathbf{v}_{\text{l}}^\text{cls} a_{\text{l}}^\text{fine} \ . 
\end{align}
We compare the coarse scores $a_{\text{l}}^\text{coarse}$ and fine-grained scores $a_{\text{l}}^\text{fine}$ in Fig.~\ref{fig:weight}, and find that $a_{\text{l}}^\text{fine}$ can fine-grainedly allocate larger scores than $a_{\text{l}}^\text{coarse}$ to key frames that humans perform the action peak in the non-uniform videos.

\subsection{Loss}
We fuse the coarse video embedding and fine-grained video embedding as a video embedding $\mathbf{o}$ with two feedforward layers $\text{FFN}^{\text{coarse}}(\cdot)$ and $\text{FFN}^{\text{fine}}(\cdot)$, projecting the coarse video embedding and fine-grained video embedding to the semantic of class $\text{c}$,
\begin{align}
    \mathbf{o} = \text{FFN}^\text{coarse}(\mathbf{o}^\text{coarse}) +  \text{FFN}^\text{fine}(\mathbf{o}^\text{fine}) \ ,
\end{align}
and compute the cosine similarity with the text class embedding $\mathbf{t}_{\text{c}}^{\text{cls}}$ by $\text{sim}(\mathbf{t}_{\text{c}}, \mathbf{o})$. 

We optimize our network by maximizing the similarity $y_{\text{b}, \text{c}^\text{gt}}$ between the $\text{b}\hbox{-}\text{th}$ video embedding in a batch and text embedding of its ground truth class $\text{c}^\text{gt}$, and minimizing the similarity between other video and text class embeddings $\{y_{\text{b}, \text{c}}\}_{\text{c}=1, \text{c} \neq \text{c}^\text{gt}}^{C}$.  Following \cite{oord2018representation}, we use the InfoNCE loss, 
\begin{align}
    &\mathcal{L}_{\text{T2V}} =  \frac{1}{B} \sum_{\text{b}=1}^{B} \frac{1}{\mathbf{k}_{\text{b}}} \sum_{\text{b}' \in \mathbf{k}_{\text{b}}} \log \frac{\exp(y_{\text{b}', \text{c}^\text{gt}})}{\sum_{\text{b}''=1}^{B} \exp(y_{\text{b}'', \text{c}^\text{gt}})} \ , \\
    &\mathcal{L}_{\text{V2T}} =  \frac{1}{B} \sum_{\text{b}=1}^{B} \frac{1}{\mathbf{k}_{\text{b}}} \sum_{\text{b}' \in \mathbf{k}_{\text{b}}} \log \frac{\exp(y_{\text{b}', \text{c}^\text{gt}})}{\sum_{\text{c}=1}^{C} \exp(y_{\text{b}', \text{c}})} \ ,\\
    &\mathcal{L} = \mathcal{L}_{\text{T2V}} + \lambda  \mathcal{L}_{\text{V2T}},
\end{align}
where $B$ is the number of batches, $\mathbf{k}_{\text{b}}$ find the index of the video that has the same class with the $\text{b}\hbox{-}\text{th}$ video, and $\lambda$ is a hyperparameter.

\begin{table}[t]
    \caption{\it \small Comparisons for multi-label action recognition on the Charades dataset \cite{sigurdsson2016hollywood}.
    }
    
    \centering
    \setlength{\tabcolsep}{5.0pt}
    \resizebox{\linewidth}{!}{%
    \begin{tabular}{lcc}
    	\toprule
    	\textbf{Method}  & \textbf{Frames}  &\multicolumn{1}{p{1.6cm}}{\centering \textbf{mAP}} \\ 
        \midrule
        MultiScale TRN \cite{zhou2018temporal} & - & 25.2\\
        STM \cite{jiang2019stm} & 16 & 35.3\\
        SlowFast+NL \cite{feichtenhofer2019slowfast} & 16+64 & 42.5 \\
        X3D-XL(312) \cite{feichtenhofer2020x3d} & 16 & 43.4 \\
        ActionCLIP~\cite{wang2021actionclip} & 32 & 44.3 \\ 
        BIKE \cite{wu2023bidirectional} & 16 & 50.4\\
        \midrule
        \textbf{Ours} & 16 &  \textbf{51.1}\\
    	\bottomrule
    \end{tabular}}
    \label{tab:charades}  
\end{table}

\begin{table}[t]
    \caption{\it \small Comparisons of zero-shot action recognition on the HMDB-51 \cite{kuehne2011hmdb} and UCF-101 \cite{soomro2012dataset} datasets. The ``VZ'' column denotes if the method is developed for zero-shot action recognition or adapted from CLIP.
    }
    \centering
    \setlength{\tabcolsep}{5.0pt}
    \resizebox{\linewidth}{!}{%
    \begin{tabular}{lccc}
    	\toprule
    	\textbf{Method}  & \textbf{VZ}  &\multicolumn{1}{p{1.6cm}}{\centering \textbf{UCF-101}} & \multicolumn{1}{p{1.6cm}}{\centering \textbf{HMDB-51}} \\ 
        \midrule
         E2E\cite{brattoli2020rethinking} & \cmark & 44.1 & 29.8 \\
         ER\cite{chen2021elaborative}& \cmark & 51.8 & 35.3  \\
         ResT\cite{lin2022cross} & \cmark & 58.7 & 41.1 \\
        X-CLIP \cite{ni2022expanding} & \xmark & 72.0  & 44.6 \\
        DIST \cite{qing2023disentangling} &\xmark & 72.3& 55.4\\
        Vita-CLIP \cite{wasim2023vita} & \xmark & 75.0 & 48.6\\
        BIKE \cite{wu2023bidirectional} & \xmark & 78.4  & 55.6\\
        M2-CLIP \cite{wang2024multimodal} & \xmark & 78.7 & 47.1\\
        \midrule
        {\textbf{Ours}} & \xmark &  \textbf{78.9}  & \textbf{56.6} \\
    	\bottomrule
    \end{tabular}}
    \label{tab:sota_zero}  
\end{table}

\begin{table}[t]
    \caption{\it \small Comparisons on few-shot action recognition across the HMDB-51 \cite{kuehne2011hmdb}, UCF-101 \cite{soomro2012dataset} and Kinetics-400 datasets \cite{kay2017kinetics}.}
    \centering
    \setlength{\tabcolsep}{1.0pt}
    \resizebox{\linewidth}{!}{%
    \begin{tabular}{lcccc}
        \toprule
        \textbf{Method}  & \textbf{Shot}  &\textbf{HMDB-51} &\textbf{UCF-101} &\textbf{Kinetics-400}\\ 
        \midrule
        VideoSwin \cite{zhou2018temporal} & 2 &20.9 &53.3& -\\
        ActionCLIP~\cite{wang2021actionclip} & 2 &54.5 &80 & 60.5\\
        BIKE \cite{wu2023bidirectional} & 2 & 65.0 &91.3& 72.9\\
        \midrule
        \textbf{Ours} & 2 &\textbf{66.0} & \textbf{92.0} &\textbf{73.8}\\
        \bottomrule
    \end{tabular}}
    
    \label{tab:fewshot}  
\end{table}

\section{Experiment}
\label{Experiment}
\subsection{Experimental Setup}
Our proposed sub-text set and code will be made available online for future studies and comparisons.
\label{sec:experimental details}

\vspace{+0.5mm}
\noindent{\bf Datasets.} We experiment across four extensively recognized video benchmarks: Kinetics-400~\cite{kay2017kinetics}, Charades \cite{sigurdsson2016hollywood}, UCF-101~\cite{soomro2012dataset}, and HMDB-51~\cite{kuehne2011hmdb} datasets. 

\vspace{+0.5mm}
\noindent{\bf Supervised Learning.} 
Our model is implemented using the PyTorch framework. We train our network with batch size 256 for 30 epochs using the AdamW optimizer. The learning rate is set to $5 \times 10^{-5}$, and we use the cosine annealing strategy with 5 warm-up epochs. We follow \cite{wu2023bidirectional} for data augmentation in training.

\vspace{+0.5mm}
\noindent{\bf Zero-shot Learning.} We evaluate our model, pre-trained on Kinetics-400~\cite{kay2017kinetics}, using the UCF-101~\cite{soomro2012dataset} and HMDB-51~\cite{kuehne2011hmdb}.

\vspace{+0.5mm}
\noindent{\bf Few-shot Learning.} We follow \cite{wang2021actionclip} to select 2 few-shot examples per human action category for training. We train our network for 2 epochs and use the same settings as in the supervised learning. Importantly, we do not use the model pre-trained on the Kinetics-400 dataset for few-shot learning on other datasets.


\vspace{+0.5mm}
\noindent{\bf Evaluation Metrics.}
We evaluate our model with top-1 and top-5 accuracy on the single-label datasets. For the multi-label dataset Charades, we follow \cite{jiang2019stm} to report mean average precision (mAP).


\subsection{Main Results}
\label{sec:main results}

\noindent{\bf Action Recognition.} We compare our network against state-of-the-art methods that utilize large-scale visual pre-training and visual-text pre-training on the Kinetics-400 dataset in Tab.~\ref{tab:k400_sota}. All methods use the ViT-B/16 backbone. 

Using fewer input frames and views during testing, our method that employs only an 8-frame input evaluated with a single view achieves the highest top-1 and top-5 accuracy. For example, the second-best method of large-scale visual-text pre-training uses 4 temporal clips on 3 spatial crops of a video but achieves top-1 accuracy 0.6\% lower than ours.

\vspace{+0.5mm}
\noindent{\bf Multi-Label Action Recognition.} We evaluate our method for multi-label action recognition on the Charades dataset in Tab. \ref{tab:charades}. The mAPs of state-of-the-art methods are reported, and we follow \cite{feichtenhofer2020x3d} to use 16 frames. Our method finds 51.1 mAP, which is a 0.7 mAP improvement over the second-best method.

\vspace{+0.5mm}
\noindent{\bf Zero-shot Action Recognition.} Our method is trained with supervision from texts and can be used for zero-shot action recognition. We compare our network with methods that adapt from CLIP, which is pre-trained on images, and with zero-shot methods developed for videos on the UCF-101 and HMDB-51 datasets in Tab.~\ref{tab:sota_zero}. All methods use a single view during testing.

Our method demonstrates superior generalization capabilities in zero-shot action recognition. Specifically, compared to the latest zero-shot methods developed for video~\cite{wang2024multimodal}, our zero-shot performance on the HMDB51 dataset is 9.5\% higher than 47.1\%. This further demonstrates the versatility of our pipeline.

\vspace{-0.1cm}
\paragraph{Few-shot Action Recognition.} We explore two-shot action recognition on the HMDB-51 \cite{kuehne2011hmdb}, UCF-101 \cite{soomro2012dataset}, and Kinetics-400 datasets \cite{kay2017kinetics} in Tab.~\ref{tab:fewshot}. With a limited amount of videos, our approach that decomposes a video action into atomic actions exhibits the highest performance. 

\begin{figure}[t]
    \centering
    \begin{subfigure}[b]{0.225\textwidth}
        \centering
        \begin{tikzpicture}
            \begin{axis}[
                grid,
                ybar,
                bar width=.371cm,
                tick align=inside,
                xlabel={Number of sub-texts ($N$)},
                ylabel={Top-1 Acc (\%)},
                width=1.0\linewidth, 
                height=1.0\linewidth, 
                xtick=data,
                yticklabels={84.4, 84.5, 84.6},
                ymin=84.35, ymax=84.6,
                xmin=1.7, xmax=5.3,
                ytick={84.4, 84.5, 84.6},
                ylabel style={yshift=-.7em},
                xlabel style={yshift=0.3em},
                enlarge x limits=0.15,
                yticklabel style={font=\small}, 
                xticklabel style={font=\small}, 
            ]
            \addplot [
                orange, 
                fill=orange!30!white
            ]
            coordinates {
                (2, 84.400) 
                (3, 84.443) 
                (4, 84.473) 
                (5, 84.490)
            };
            \end{axis}
        \end{tikzpicture}
        \caption{The impact of numbers of sub-texts ($N$).}
        \label{fig:ablation:num_sub-texts}
    \end{subfigure}
    \hfill
    \begin{subfigure}[b]{0.225\textwidth}
        \centering
        \raisebox{-0.0em}{
        \begin{tikzpicture}
            \begin{axis}[
                grid,
                xlabel={\small $\text{TPP}$ },
                ylabel={Top-1 Acc (\%)},
                legend style={at={(0.98,0.98)}, anchor=north east, font=\tiny, draw=none, fill=none},
                width=1.05\linewidth, 
                height=1.0\linewidth, 
                yticklabels={84.4, 84.5},
                xmin=25, xmax=65,
                xtick distance = 50,
                ymin=84.35, ymax=84.5,
                ytick={84.4, 84.5},
                xtick={25, 35, 45, 55, 65},
                enlarge x limits=0.15,
                ylabel style={yshift=-2.0em},
                xlabel style={yshift=0.3em},
                yticklabel style={font=\small}, 
                xticklabel style={font=\small}, 
                label style={font=\tiny}, 
                y label style={at={(axis description cs:-0.1,.5)},anchor=south, font=\small},
            ]
            \addplot[only marks, mark=*] coordinates {
                (28.52, 84.37) 
                (34.49, 84.41) 
                (38.18, 84.44)
                (41.48, 84.43)
                (43.89, 84.43)
                (45.30, 84.45)
                (46.10, 84.45) 
                (47.47, 84.45) 
                (49.64, 84.46) 
                (54.68, 84.45) 
                (60.43, 84.47) 
            };
            \addplot[dashed, color=orange, line width=1pt] coordinates {
            (25, 84.3836)
            (63, 84.4932)
            };
            \end{axis}
        \end{tikzpicture}}
        \vspace{-3mm}
        \caption{Study on the relationship between $\text{TPP}$ and Top-1.}
        \label{fig:ablation:TPP_performance}
    \end{subfigure}

    \caption{\it \small Analysis of sub-texts. We study (a) the impact of the number of sub-texts on action recognition performance, and (b) the correlation of \textnormal{TPP} (average of \textnormal{$\text{TPP}_{\text{c}}$} for all actions) and action recognition performance, with the fitted dashed line showing an \textnormal{$r^2$} value of 0.79.
    }
    \label{fig:combined_analysis}
    \vspace{-4mm}
\end{figure}
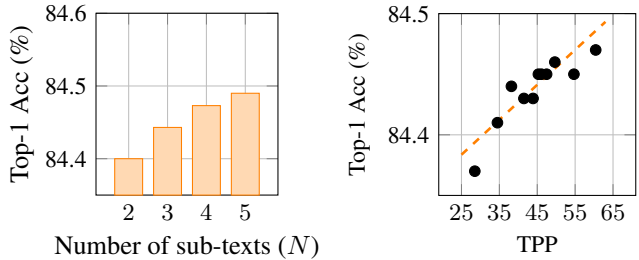

\subsection{Ablation Studies and Analysis}
\label{sec:Ablation Studies}
We perform ablation studies with the Kinetics-400 dataset to examine our approach.


\vspace{+0.5mm}
\noindent{\bf Number of Sub-texts.} Fig.~\ref{fig:ablation:num_sub-texts} illustrates the number ${N}$ of sub-texts used in our model. As the number of sub-texts increases from 2 to 5, the recognition accuracy of our method improves. However, we observe that the performance gain is minimal when increasing the number of sub-texts from 4 to 5. Considering both computational complexity and recognition accuracy, we opted for 4 sub-texts in our experiments. 

\begin{table}[t]
    \caption{\it \small Ablations of coarse and fine-grained video embedding (VE). We use fine-tuned CLIP model and CLIP model with temporal layers as our baselines.}
    \centering
    \resizebox{\linewidth}{!}{%
        \begin{tabular}{cc|cc}
            \toprule
             \textbf{Coarse VE} &  \textbf{Fine-grained VE} & \textbf{Top-1(\%)} &\textbf{Top-5(\%)} \\ \midrule
            \multicolumn{2}{c|}{Baseline (CLIP)} & 79.9 & 94.7 \\
             \multicolumn{2}{c|}{Baseline (Temporal)} & 80.3 & 95.0 \\
             \xmark & \cmark & 82.7 & 96.2 \\
            \cmark & \xmark & 83.1 & 96.0 \\
            \cmark & \cmark & \textbf{84.5} & \textbf{96.7} \\
            \bottomrule
        \end{tabular}}
    \label{tab:video}
    \vspace{-4mm}
\end{table}

\begin{table}[t]
    \caption{\it \small Generalization of our components (OC) of coarse video embedding and fine-grained video embedding on state-of-the-art action recognition methods.}
    \centering
    \setlength{\tabcolsep}{5pt}
    \resizebox{\linewidth}{!}{%
        \begin{tabular}{lccc}
        \toprule
        \textbf{Backbone} &  \textbf{OC} &  \textbf{Top-1(\%)} &  \textbf{Top-5(\%)} \\ 
            \midrule
            \multirow{2}{*}{VideoPrompt~\cite{ju2022prompting}} & \textcolor{black}{\xmark} & 76.9 & 93.5 \\ 
            &\textcolor{black}{\cmark}& 79.3& 95.1\\  
            \midrule
            \multirow{2}{*}{ATM~\cite{atm}} & \textcolor{black}{\xmark}  & 82.8 & 95.6 \\ 
            &\textcolor{black}{\cmark}& 82.9& 96.4\\ 
            \midrule
            \multirow{2}{*}{BIKE~\cite{wu2023bidirectional}} & \textcolor{black}{\xmark}  & 83.2 & 96.0 \\ 
            &\textcolor{black}{\cmark}& 83.8& 96.5\\ 
        \bottomrule
        \end{tabular}}
    \label{tab:othermethods}
    \vspace{-4mm}
\end{table}	

\noindent{\bf Correlation of $\text{TPP}$ and Performance.}
In Fig.~\ref{fig:ablation:TPP_performance}, 11 sub-text groups are generated, and we show the relationship between $\text{TPP}$ and the action recognition performance, where $\text{TPP}$ is an average of $\text{TPP}_{\text{c}}$ for all actions. We observe a positive correlation with an $r^2$ value of 0.79, demonstrating the effectiveness of our $\text{TPP}$ method in selecting sub-texts without the need for computational experiments.


\vspace{+0.5mm}
\noindent{\bf Effectiveness of Components.} We study the effective usage of our coarse video embedding and fine-grained video embedding in Tab. \ref{tab:video}. A CLIP model is fine-tuned on the Kinetics-400 dataset to establish a performance baseline. Following the common practice of existing works, we build a baseline (Temporal) using a 6-layer Transformer encoder. Our coarse and fine-grained video embeddings consistently improve upon this baseline. For example, using all embeddings, we achieve a top-1 accuracy of 84.5\%, which is 4.6\% higher than the baseline.


\vspace{+0.5mm}
\noindent{\bf Generalization.} To validate the generalization of our components of coarse video embedding and fine-grained video embedding, we apply them to state-of-the-art action recognition methods, as shown in Tab.~\ref{tab:othermethods}. Specifically, we study VideoPrompt \cite{ju2022prompting}, ATM \cite{atm} and BIKE~\cite{wu2023bidirectional}. The results indicate that we enhance the accuracy of these state-of-the-art methods, such as VideoPrompt, which shows a 2.4\% increase in accuracy on the Kinetics-400 dataset.

\vspace{+0.5mm}
\noindent{\bf More.} We provide additional implementation details, comparisons, and ablation studies in the supplementary material.


\section{Conclusion}
\label{Conclusion}
In this paper, we propose a framework to transfer CLIP trained on image-text pairs to video action recognition. Similar to how a video forms a video action by performing a sequence of atomic actions, our key insight is to decompose a video action into a sequence of atomic action descriptions using a pre-trained LLM. We then select these atomic action descriptions with a proposed metric. The global and atomic action descriptions are used to identify salient video frames from ambiguous and non-uniform videos for action recognition. Experiments on standard benchmark datasets demonstrate that our method significantly outperforms previous works in supervised, few-shot, and zero-shot settings.

\bibliography{aaai25}

\end{document}